%% file: main.tex
\documentclass[runningheads]{llncs}
\usepackage{graphicx}
\usepackage{comment}
\usepackage{amsmath,amssymb, amsfonts}
\usepackage{color}
\usepackage{graphicx}
\usepackage{booktabs}
\usepackage{caption}

\usepackage[colorlinks, citecolor=cyan]{hyperref}
\usepackage[dvipsnames]{xcolor}
\usepackage[mathscr]{euscript}
\usepackage{nicefrac}
\usepackage{bbm}
\usepackage{microtype}
\usepackage{multirow}
\usepackage{tabularx}
\usepackage{subcaption}

\newcommand{\R}{\mathbb{R}}

\newcommand{\loss}[1]{L_\text{#1}}
\newcommand{\norm}[1]{\lVert{#1}\rVert}

\newcommand{\eref}[1]{(\ref{#1})}
\newcommand{\secref}[1]{Sec.~\ref{#1}}
\newcommand{\figref}[1]{Fig.~\ref{#1}}
\newcommand{\tabref}[1]{Tab.~\ref{#1}}

\makeatletter
\def\blfootnote{\xdef\@thefnmark{}\@footnotetext}
\makeatother

\begin{document}
\pagestyle{headings}
\mainmatter
\title{Perceiving 3D Human-Object Spatial Arrangements from a Single Image in the Wild}

\titlerunning{PHOSA: Perceiving Human-Object Spatial Arrangements}
\author{Jason Y. Zhang\index{test}\inst{1}* \and
Sam Pepose\inst{2}* \and
Hanbyul Joo\inst{2} \and\\
Deva Ramanan\inst{1,3} \and
Jitendra Malik\inst{2,4} \and
Angjoo Kanazawa\inst{4}
}
\authorrunning{J.Y. Zhang et al.}
\institute{{$^1$ Carnegie Mellon University} \hspace{15mm} {$^2$ Facebook AI Research} \hspace{15mm} \\ \hspace{10mm}{$^3$ Argo AI}  \hspace{35mm} {$^4$ UC Berkeley} \hspace{10mm}}
\maketitle

\begin{abstract}
\input{00_abstract}
\end{abstract}
\begin{figure}
    \centering
    \includegraphics[width=\textwidth]{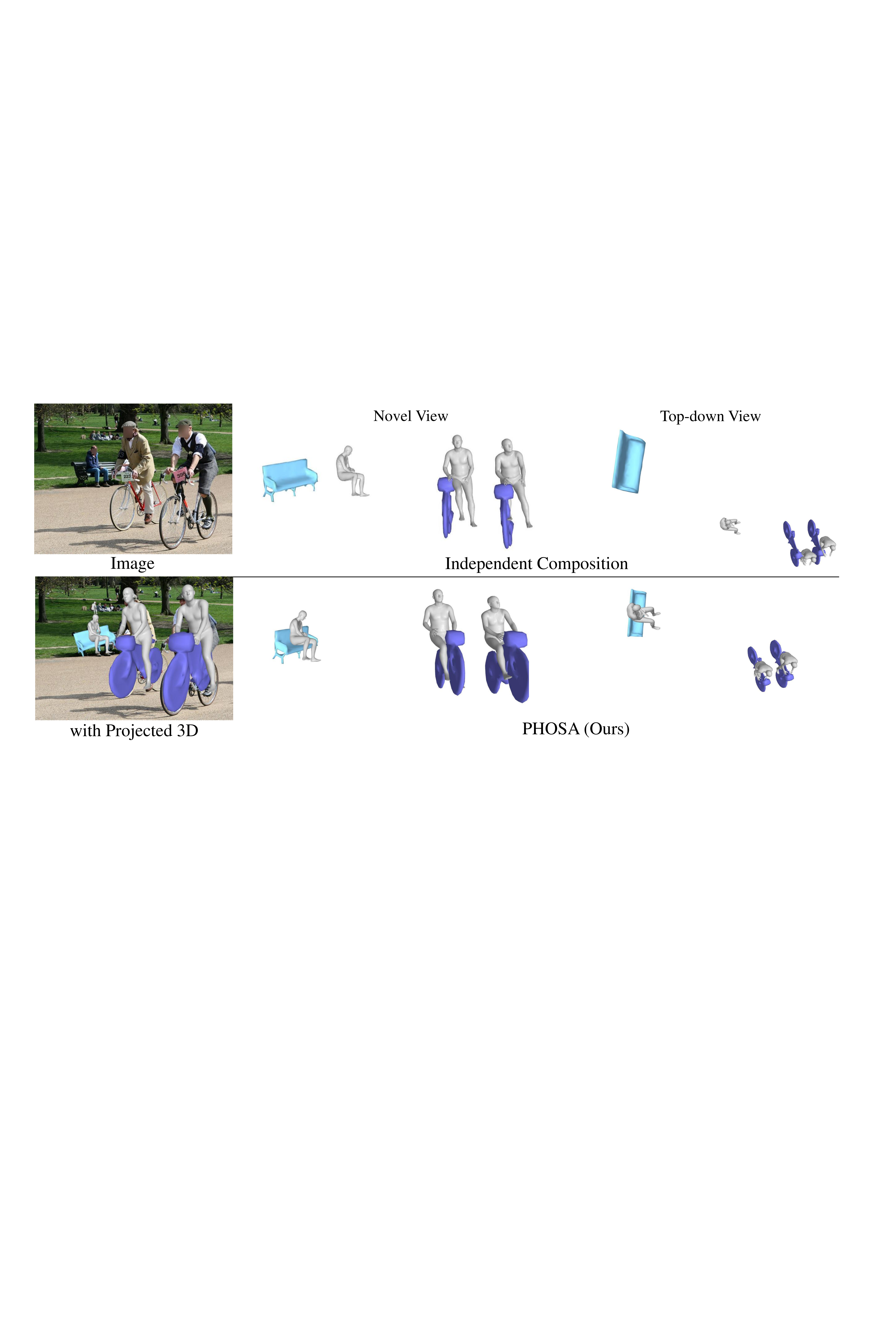}
    \caption{We present PHOSA, Perceiving Human-Object Spatial Arrangements, an approach that recovers the spatial arrangements of humans and objects in 3D space from a single image by reasoning about their intrinsic scale, human-object interaction, and depth ordering. 
    Given the input image (top left), we show two possible interpretations of the 3D scene that have similar 2D projections (bottom left). Using priors of humans, objects, and their interactions, our approach is able to recover the more reasonable interpretation (bottom row).}
    \label{fig:teaser}
\end{figure}
\blfootnote{* denotes equal contribution. Please direct correspondence to \texttt{jasonyzhang@cmu.edu}.}
\input{01_introduction}
\input{02_related}

\input{03_methods}

\input{04_experiments}
\input{05_discussion}

\bibliographystyle{splncs04}
\bibliography{egbib}

\clearpage
\input{06_appendix}

\end{document}

%% file: 00_abstract.tex
We present a method that infers spatial arrangements and shapes of humans and objects in a globally consistent 3D scene, all from a single image in-the-wild captured in an uncontrolled environment.
Notably, our method runs on datasets without any scene- or object-level 3D supervision. 
Our key insight is that considering humans and objects jointly gives rise to ``3D common sense" constraints that can be used to resolve ambiguity.
In particular, we introduce a scale loss that learns the distribution of object size from data;
an occlusion-aware silhouette re-projection loss to optimize object pose; and a human-object interaction loss to capture the spatial layout of objects with which humans interact.
We empirically validate that our constraints dramatically reduce the space of likely 3D spatial configurations.
We demonstrate our approach on challenging, in-the-wild images of humans interacting with large objects (such as bicycles, motorcycles, and surfboards) and handheld objects (such as laptops, tennis rackets, and skateboards). 
We quantify the ability of our approach to recover human-object arrangements and outline remaining challenges in this relatively unexplored domain. The project webpage can be found at \url{https://jasonyzhang.com/phosa}.

%% file: 01_introduction.tex
\section{Introduction}

Tremendous strides have been made in estimating the 2D structure of in-the-wild scenes in terms of their constituent objects. 
While recent work has also demonstrated impressive results in estimating 3D structures, particularly human bodies, the focus is often on bodies~\cite{hmr,kolotouros2019spin} and objects~\cite{choy20163d,girdhar16b} imaged in isolation or in controlled lab conditions~\cite{prox,PiGraph}.
To enable true 3D in-the-wild scene understanding, we argue that one must look at the {\em holistic} 3D scene, where objects and bodies can provide contextual cues for each other so as to correct local ambiguities. Consider the task of understanding the image in Figure \ref{fig:teaser}.
Independently estimated 3D poses of humans and objects are not necessarily consistent in the spatial arrangement of the 3D world of the scene (top row).
When processed holistically, one can produce far more plausible 3D arrangements by exploiting contextual cues, such as the fact that 
humans tend to sit on park benches and ride bicycles rather than float mid-air.

In this paper, we present a method that can similarly recover the 3D spatial arrangement and shape of humans and objects in the scene from a single image.
We demonstrate our approach on challenging, in-the-wild images containing multiple and diverse human-object interactions. We propose an optimization framework that relies on automatically predicted 2D segmentation masks to recover the 3D pose, shape, and location of humans along with the 6-DoF pose and intrinsic scale of key objects in the scene.
Per-instance intrinsic scale allows one to convert each instance's local 3D coordinate system to coherent world coordinates, imbuing people and objects with a consistent notion of metric size.

There are three significant challenges to address. First is that the problem is inherently ill-posed as multiple 3D configurations can
result in the same 2D projection. It is attractive to make use of data-driven priors to resolve such ambiguities. But we immediately run into the second challenge: obtaining training data with 3D supervision is notoriously challenging, particularly for entire 3D scenes captured in-the-wild. Our key insight is that considering humans and objects jointly gives rise to 3D scene constraints that reduce ambiguity.
We make use of physical 3D constraints including a prior on the typical size of objects within a category. We also incorporate spatial constraints that encode typical modes of interactions with humans (e.g.~humans typically interact with a bicycle by grabbing its handlebars).
Our final challenge is that while there exists numerous mature technologies supporting 3D understanding of humans (including shape models and keypoint detectors), the same tools do not exist for the collective space of all objects.
In this paper, we take the first step toward building such tools by learning the natural size distributions of object categories without any supervision.
Our underlying thesis, bourne out by experiment, is that contextual cues arising from holistic processing of human-object arrangements can still provide enough information to understand objects in 3D.

We design an optimization-based framework, where we first reconstruct the humans and objects {\em locally} in each detected bounding box. For humans, we make use of state-of-the-art 3D human reconstruction output~\cite{joo2020exemplar}. For objects, we solve for the 6-DoF parameters of a category-specific 3D shape exemplar that fits the local 2D object instance segmentation mask \cite{kato2018nmr}. We then use a per-instance intrinsic scale to convert each local 3D prediction into a world coordinate frame by endowing metric size to each object and define a {\em global} objective function that scores different 3D object layouts, orientations, and shape exemplars. We operationalize constraints through loss terms in this objective. 
We make use of gradient-based solvers to optimize for the globally consistent layout.
Although no ground truth is available for this task, we evaluate our approach qualitatively and quantitatively on the COCO-2017 dataset \cite{lin2014coco}, which contains challenging images of humans interacting with everyday objects obtained in uncontrolled settings.
We demonstrate the genericity of our approach by evaluating on objects from 8 categories of varying size and interaction types: baseball bats, bicycles, laptops, motorcycles, park benches, skateboards, surfboards, and tennis rackets.

%% file: 02_related.tex
\section{Related Work}

\begin{figure*}[t]
	\centering
	
	\includegraphics[width=\textwidth]{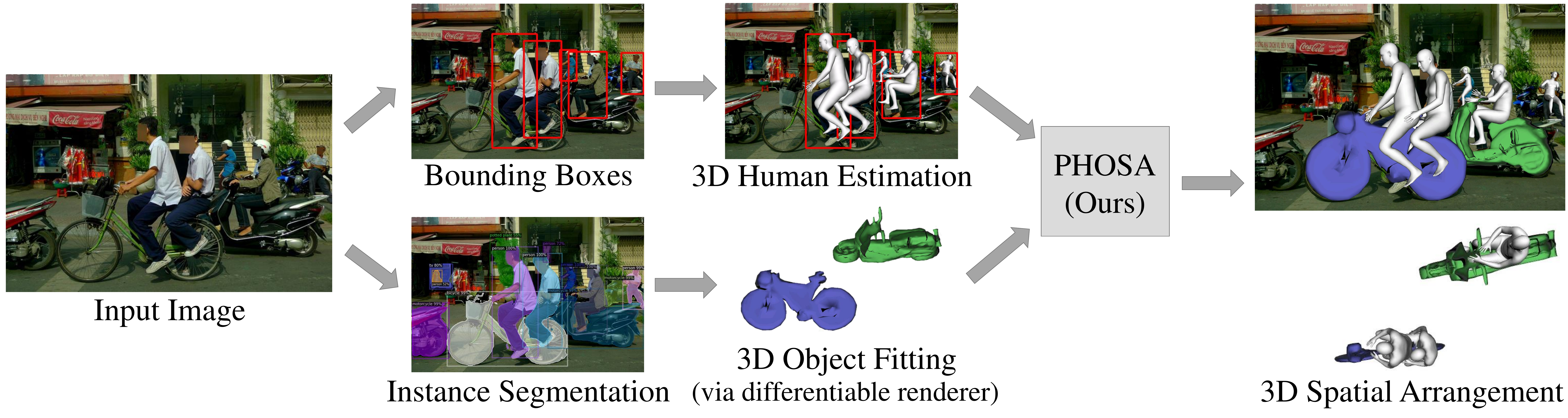}
	\caption{\textbf{Overview of our method, PHOSA.} Given an image, we first detect instances of humans and objects~\cite{maskRCNN}. We predict the 3D pose and shape of each person~\cite{joo2020exemplar} and optimize for the 3D pose of each object by fitting to a segmentation mask~\cite{kato2018nmr}. 
		Then, we convert each 3D instance in its own local coordinate frame into world coordinates using an intrinsic scale. Using our Human-Object Spatial Arrangement optimization, we produce a globally consistent output, as shown here.
		Our framework produces plausible reconstructions that capture realistic human-object interaction, preserve depth ordering, and obey physical constraints.
	}
	\label{fig:Overview}
\end{figure*}

\paragraph{\textbf{3D human pose and shape from a single image.}}
Recovering the 3D pose and shape of a person from a single image is a fundamentally
ambiguous task. As such, most methods employ statistical 3D body models with strong priors on shape learned from large-scale 3D scans and with known kinematic structure to model the articulation \cite{Anguelov:2005,zhou2010parametric,SMPL,SMPL-X:2019,joo2018}. 
Seminal works in this area
\cite{grauman2003inferring,sigal2008combined,agarwal2006recovering} fit the parameters of a 3D body model to manually annotated silhouettes or keypoints \cite{sigal2008combined,guan2009estimating,zhou2010parametric}. Taking advantage of the progress in 2D pose estimation, \cite{SMPLify} propose a fully automatic approach where the parameters of the SMPL body model \cite{SMPL} are fit to automatically detected 2D joint locations in combination with shape and pose priors. More recent approaches employ expressive human models with faces and fingers \cite{xiang19monocular,pavlakos2019smlx}. 
Another line of work develops a learning-based framework, using a feed-forward model to directly predict the parameters of the body model from a single image~\cite{Tung2017,pavlakos2018humanshape,hmr,omran2018nbf,varol2018bodynet,VNect_SIGGRAPH2017}. 
\cite{guler2019holopose} combine human detection with 3D human pose and shape prediction. Much of the focus in such approaches is training on in-the-wild images of humans without paired 3D supervision. \cite{hmr} employ an adversarial prior on pose, \cite{pavlakos2018ordinal} explore using ordinal supervision, and \cite{pavlakos2019texturepose} use texture consistency. 
To achieve state-of-the-art results, \cite{kolotouros2019spin,joo2020exemplar} have proposed hybrid approaches that combine feed-forward networks with optimization-based methods to improve 2D keypoint fit. In this work, we use the 3D regression network from \cite{joo2020exemplar} to recover the 3d pose and shape of humans.

Note that all of these approaches consider the human in isolation. More related to our work are methods that recover 3D pose and shape of multiple people \cite{zanfir18monocular,zanfir18deep,jiang2020coherent}. These approaches use collision constraints to avoid intersection, and use bottom-up grouping or ordinal-depth constraints to resolve ambiguities.
We take inspiration from these works for the collision loss and the depth ordering loss from \cite{jiang2020coherent}, but our focus is on humans and objects in this work.

\paragraph{\textbf{3D objects from a single image.}}
There has also been significant literature in single-view 3D object reconstruction. Earlier methods optimize a deformable shape model to image silhouettes \cite{lim2013parsing,kholgade20143d,aubry2014seeing,CSDM,izadinia2017im2cad}. Recent approaches train a deep network to predict the 3D shape from an image \cite{choy20163d,girdhar16b,psgn,groueix2018papier,park2019deepsdf,mescheder2019occupancy,chen2019learning}. Most of these approaches require 3D supervision or multi-view cues and are trained on synthetic datasets such as \cite{wu20153d,song2017semantic}. Many of these approaches reason about 3D object shape in isolation, while in this work we focus on their spatial arrangements. There are several works \cite{factored3dTulsiani17,kundu20183d,meshrcnn} that recover the 3D shape of multiple objects but still reason about them independently. More recently, \cite{kulkarni20193d} proposes a graph neural network to reason about the relationships between object to infer their layout, trained on a synthetic dataset with no humans. In this work we explore 3D spatial arrangements of humans and objects in the wild. As there is no 3D supervision for these images, we take the traditional category-based model-fitting approach to get the initial 6-DoF pose of the objects and refine their spatial arrangements in relation to humans and other objects in the scene.

\paragraph{\textbf{3D human-to-object interaction.}}

Related to our work are earlier approaches that infer about the 3D geometry and affordances from observing people interacting with scenes over time \cite{gupta20113d,delaitre2012scene,fouhey2014people,li2019estimating}.
These approaches are similar to our work in spirit in that they use the ability to perceive humans to understand the 3D properties of the scene. 
The majority of the recent works rely on a pre-captured 3D scene to reason about 3D human-object interaction. \cite{PiGraph}
use RGB-D sensors to capture videos of people interacting with indoor scenes and use this data to learn a probabilistic model that reasons about how humans interact with its environment. Having access to 3D scenes provides scene constraints that improve 3D human pose perception \cite{yamamoto2000scene,rosenhahn2008markerless,kjellstrom2010tracking}. 
The PROX system \cite{prox} demonstrates this idea through an optimization-based approach to improve 3D human pose estimation conditioned on a known 3D scene captured by RGB-D sensors. While we draw inspiration from their contact terms to model human-object interaction, we critically do not assume that 3D scenes are available.
\cite{Rosinol-RSS-20} is another recent approach that constructs a 3D scene graph of humans and objects from video and inertial data captured indoors.
We, on the other hand, experiment on single images captured in uncontrolled {\em in-the-wild} environments, often outdoors.

More related to our approach are those that operate on images. There are several hand-object papers that recover both 3D object and 3D hand configurations \cite{hasson2019learning}. In this work we focus on 3D spatial arrangements of humans and objects. 
Imapper~\cite{monszpart2019imapper} uses priors built from RGB-D data \cite{PiGraph} to recover a plausible global 3D human motion and a global scene layout from an indoor video. 
Most related to our work is \cite{chen2019holistic++}, who develop an approach that recovers a parse graph that represents the 3D human pose, 3D object, and scene layout from a single image. They similarly recover the spatial arrangements of humans and objects, but rely on synthetic 3D data to (a) learn priors over human-object interaction and (b) train 3D bounding box detectors that initialize the object locations. In this work we focus on recovering 3D spatial arrangements of humans and objects in the wild where no 3D supervision is available for 3D objects and humans and their layout. Due to the reliance on 3D scene capture and/or 3D synthetic data, many previous work on 3D human object interaction focus on indoor office scenes. By stepping out of this supervised realm, we are able to explore and analyze how 3D humans and objects interact in the wild. 

%% file: 03_methods.tex
\section{Method}

Our method takes a single RGB image as input and outputs humans and various categories of objects in a common 3D coordinate system. We begin by separately estimating 3D humans and 3D objects in each predicted bounding box provided by an object detector~\cite{maskRCNN}. 
We use a state-of-the art 3D human pose estimator~\cite{joo2020exemplar} to obtain 3D humans in the form of a parametric 3D human model (SMPL \cite{SMPL}) (in \secref{sec:anthro_rcnn}), and use a differentiable renderer to obtain 3D object pose (6-DoF translation and orientation)
by fitting 3D mesh object models to predicted 2D segmentation masks~\cite{kirillov2019pointrend} (in \secref{sec:nmr_opt}).
The core idea of our method is to exploit the interaction between humans and objects to spatially arrange them in a common 3D coordinate system by optimizing for the per-instance \textit{intrinsic scale}, which specifies their metric size (in \secref{sec:human_object}). In particular, our method can also improve the performance of 3D pose estimation for objects by exploiting cues from the estimated 3D human pose. See Fig.~\ref{fig:Overview} for the overview of our method.

\begin{figure*}[t]
	\centering
	\includegraphics[width=\textwidth]{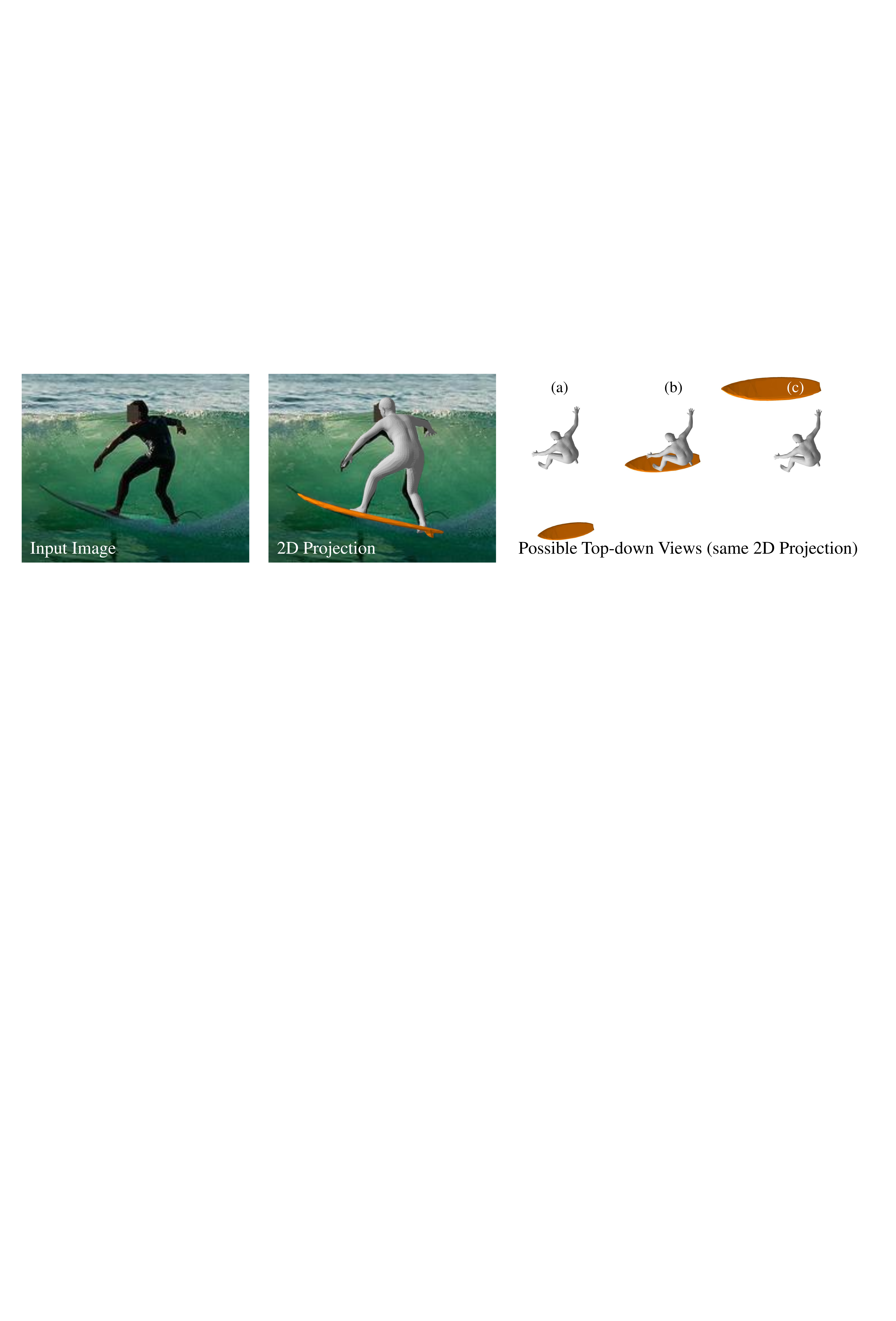}
	\caption{\textbf{Ambiguity in scale.} Recovering a 3D scene from a 2D image is fundamentally ambiguous because multiple 3D interpretations can have the same 2D projection. Consider the photo of the surfer on the left. A large surfboard far away (c) and a small one closer (a) have the same 2D projection (second panel) as the correct interpretation (b). In this work, we aim to resolve this scale ambiguity by exploiting cues from human-object interaction, essentially using the human as a ``ruler."}
	\label{fig:scale_fig}
\end{figure*}

\subsection{Estimating 3D Humans}
\label{sec:anthro_rcnn}

\begin{figure}[t]
	\centering
	
	\includegraphics[width=\textwidth]{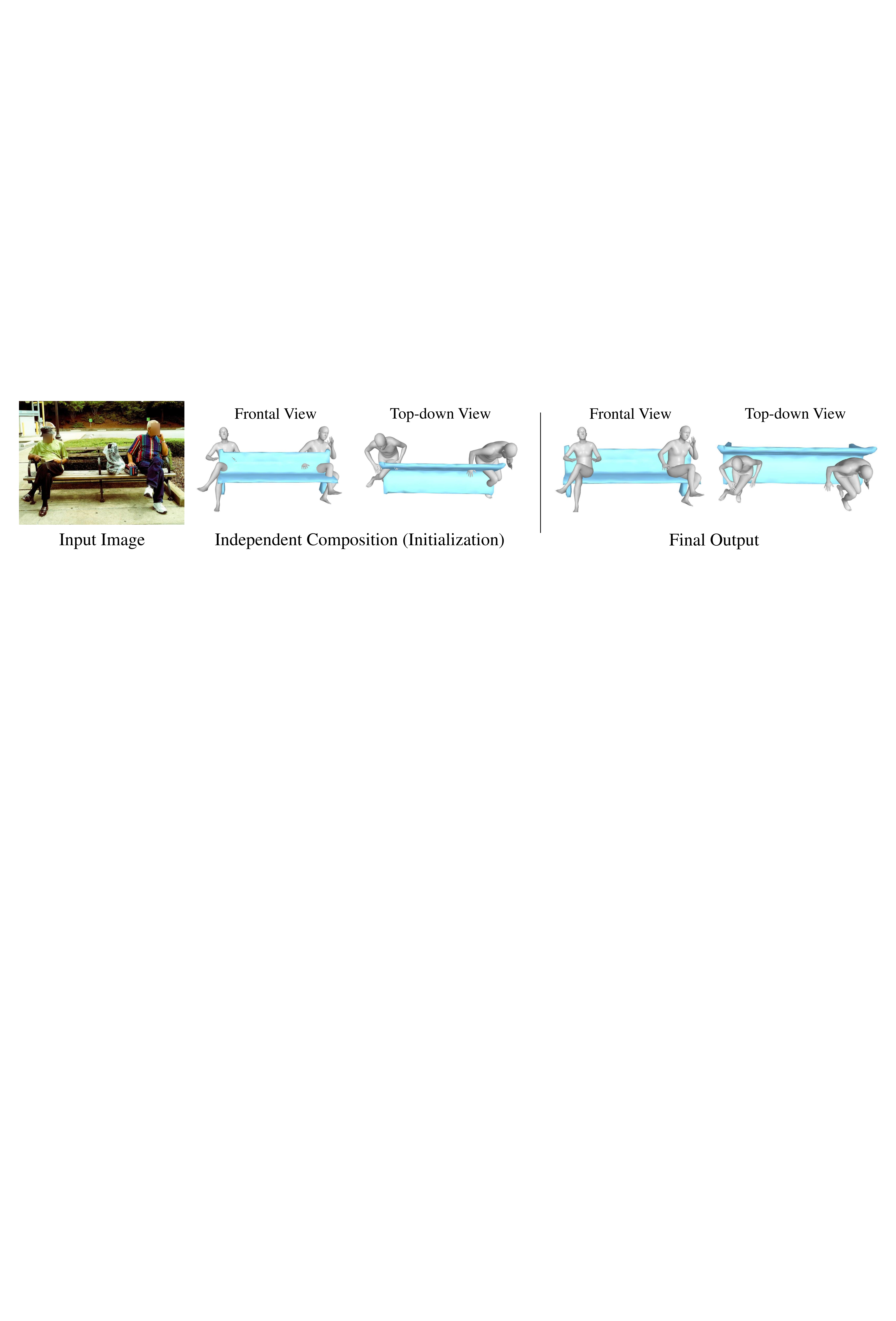}
	
	\caption{\textbf{Recovering realistic human-object spatial arrangements by reasoning about depth ordering and interpenetration.} Given the image on the left as input, we first initialize the spatial arrangement by independently estimating the 3D human and object poses (Independent Composition).
		By incorporating physical priors such as avoiding mesh inter-penetration as well as preserving the depth ordering inferred from the predicted segmentation mask, we can produce the much more plausible spatial arrangement shown on the right.}
	\label{fig:loss}
\end{figure}

Given a bounding box for a human provided by a detection algorithm~\cite{maskRCNN}, we estimate the 3D shape and pose parameters of SMPL \cite{SMPL} using~\cite{joo2020exemplar}. The 3D human is parameterized by pose $\theta \in \R^{72}$ and shape $\beta \in \R^{10}$, as well as as a weak-perspective camera $\Pi = [\sigma, t_x, t_y] \in \R^3$ to project the mesh into image coordinates. To position the humans in the 3D space, we convert the weak-perspective camera to the perspective camera projection by assuming a fixed focal length $f$ for all images, where the distance of the person is determined by the reciprocal of the camera scale parameter $\sigma$. Thus, the 3D vertices of the SMPL model for the $i$-th human is represented as:
\begin{equation}
V^i_h = \mathcal{M}(\beta^i, \theta^i) + \begin{bmatrix}t^i_x & t^i_y & \nicefrac{f}{\sigma^i}\end{bmatrix},
\label{eq:SMPL_pers}
\end{equation}
where $\mathcal{M}$ is the differentiable SMPL mapping from pose and shape to a human mesh with 6890 vertices in meters. The SMPL shape parameter $\beta$ controls the height and size of the person. In practice, this is difficult to reliably estimate from an image since a tall, far-away person and a short, closeby person may project to similar image regions (\figref{fig:teaser} and  \figref{fig:loss}).
To address this ambiguity, we fix the estimated SMPL pose and shape and introduce an additional per-human intrinsic scale parameter $s^j \in \R$ 
that changes the size and thus depth of the human in world coordinates: ${V^i_h}^* = s^j V^i_h$. While the shape parameter $\beta$ also captures size, we opt for this parameterization
as it can also be applied to objects (described below) and thus optimized to yield a globally consistent layout of the scene.

\subsection{Estimating 3D Objects}
\label{sec:nmr_opt}
\begin{figure}[t]
	\centering
	\begin{minipage}{0.49\textwidth}
		\centering
		\includegraphics[width=\textwidth]{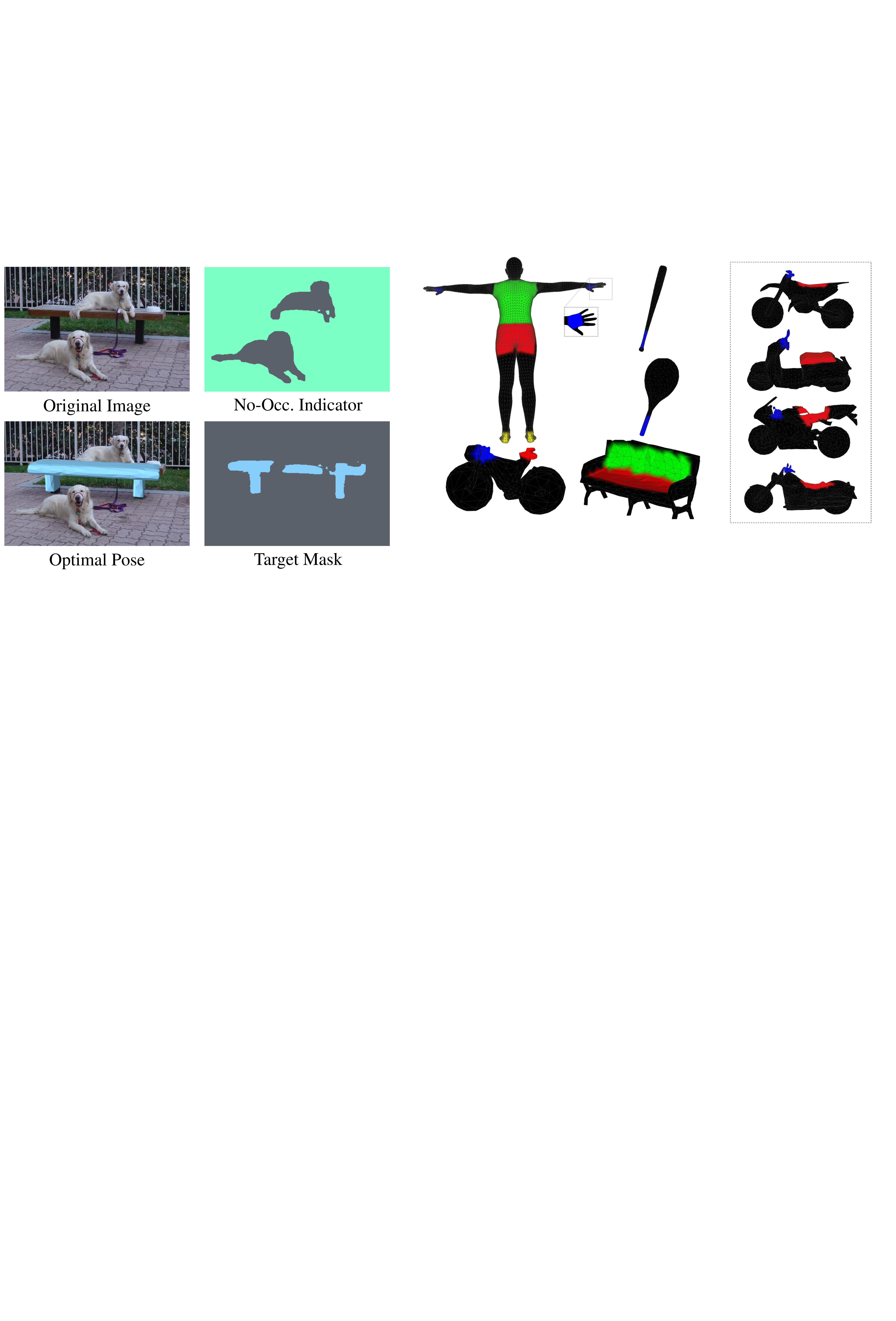}
		\caption{\textbf{Occlusion-Aware Silhouette Loss for optimizing object pose.} Given an image, a 3D mesh model, and instance masks, our occlusion-aware silhouette loss finds the 6-DoF pose that most closely matches the target mask (bottom right). To be more robust to partial occlusions, we use a no-occlusion indicator (top right) to ignore regions that correspond to other object instances, including those for which we do not have 3D mesh models (e.g.~dogs).}
		\label{fig:occ-sil-loss}
	\end{minipage}\hfill
	\begin{minipage}{0.49\textwidth}
		\centering
		\includegraphics[width=\textwidth]{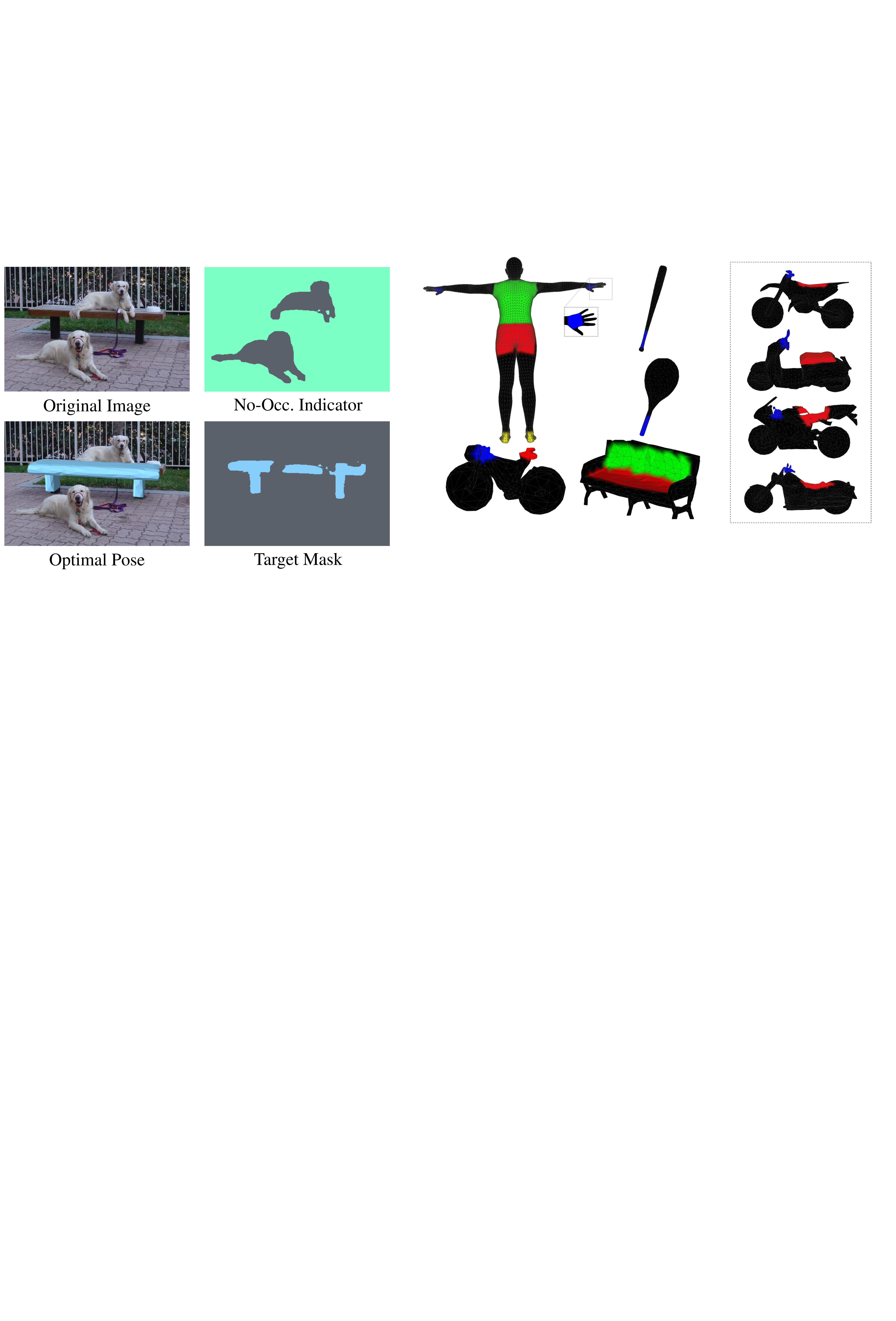}
		\caption{
			\textbf{Part labels for fine-grained interaction.}
			To model human-object interactions, we label each object mesh with interaction regions corresponding to parts of the human body. Each color-coded region on the person (top left) interacts with the matching colored region for each object. 
			The interaction loss pulls pairs of corresponding parts closer together. To better capture variation in shape, we can use multiple mesh instances for the same category (e.g.~the motorcycles shown on the right). See the supplementary in \secref{sec:supp} for all mesh models and interaction correspondences.
		}
		\label{fig:part_labels}
	\end{minipage}
\end{figure}

We consider each object as a rigid body mesh model and estimate the 3D location $\mathbf{t} \in \R^3$, 3D orientation $\mathcal{R} \in SO(3)$, and an intrinsic scale $s\in\R$. The intrinsic scale converts the local coordinate frame of the template 3D mesh to the world frame. We consider single or multiple exemplar mesh models for each object category, pre-selected based on the shape variation within each category. For example, we use a single mesh for skateboards but four meshes for motorcycle. The mesh models are obtained from \cite{warehouse,free3d,kundu20183d}
and are pre-processed to have fewer faces (about 1000) to make optimization more efficient. See \figref{fig:part_labels} for some examples of mesh models and the supplementary for a full list. The 3D state of the $j$th object is represented as:
\begin{equation}
V^j_o = s^j \left(\mathcal{R}^j \mathcal{O} ( c^j, k^j ) + \mathbf{t}^j\right),
\end{equation}
where $\mathcal{O} ( c^j, k^j )$ specifies the $k^j$-th exemplar mesh for category $c^j$. Note that the object category $c^j$ is provided by the object detection algorithm~\cite{maskRCNN}, and $k^j$ is automatically determined in our optimization framework (by selecting the exemplar that minimizes reprojection error).

Our first goal is to estimate the 3D pose of each object independently. However, estimating 3D object pose in the wild is challenging because
(1) there are no existing parametric 3D models for target objects; (2) 2D keypoint annotations or 3D pose annotations for objects in the wild images are rare; and (3) occlusions are common in cluttered scenes, particularly those with humans.
We propose an optimization-based approach using a differentiable renderer~\cite{kato2018nmr} to fit the 3D object to instance masks from~\cite{kirillov2019pointrend} in a manner that is robust to partial occlusions. We began with an pixel-wise L2 loss over rendered silhouettes $S$ versus predicted masks $M$, but found that it ignored boundary details that were important for reliable pose estimation. We added a symmetric chamfer loss \cite{gavrila2000pedestrian} which focuses on boundary alignment, but found it computationally prohibitive since it required recomputing a distance transform of $S$ at each gradient iteration. We found good results with an L2 mask loss augmented with a {\em one-way} chamfer loss that computes the distance of each silhouette boundary pixel to the nearest mask boundary pixel, which requires computing a single distance transform once for the mask $M$. Given a
no-occlusion indicator $I$ (0 if pixel only corresponds to mask of different instance, 1 else), we write our loss as follows:
\begin{equation}
\loss{occ-sil} = \sum \left(I\circ S - M\right)^2 + \sum_{p\in E(I\circ S)} \min_{\hat{p}\in E(M)} \norm{p - \hat{p}_2}
\label{loss:occ_sil}
\end{equation}
where $E(M)$ computes the edge map of mask $M$. Note that this formulation can handle partial occlusions by object categories for which we do not have 3D models, as illustrated in \figref{fig:occ-sil-loss}. We also add an offscreen penalty to avoid degenerate solutions when minimizing the chamfer loss. To estimate the 3D object pose, we minimize the occlusion-aware silhouette loss:
\begin{equation}
\{ \mathcal{R}^j, \mathbf{t}^j \}^* = \operatorname*{argmin}_{\mathcal{R}, \mathbf{t} } \loss{occ-sil} \left(  \Pi_{\text{sil}} (V^j_o), M^j  \right),
\label{eq:sil}
\end{equation}
where $\Pi_{\text{sil}}$ is the silhouette rendering of a 3D mesh model via a perspective camera with a fixed focal length (same as $f$ in \eref{eq:SMPL_pers}) and $M^j$ is a 2D instance mask for the $j$-th object. We use PointRend \cite{kirillov2019pointrend} to compute the instance masks. See \figref{fig:occ-sil-loss} for a visualization and the supplementary for more implementation details. While this per-instance optimization provides a reasonable 3D pose estimate, the mask-based 3D object pose estimation is insufficient since 
there remains a fundamental ambiguity in determining the global location relative to other objects or people, as shown in \figref{fig:scale_fig}. In other words, reasoning about instances in isolation cannot resolve ambiguity in the intrinsic scale of the object.

\subsection{Modeling Human-Object Interaction for 3D Spatial Arrangement}
\label{sec:human_object}

Reasoning about the 3D poses of humans and objects independently may produce inconsistent 3D scene arrangements.
In particular, objects suffer from a fundamental depth ambiguity: a large object further away can project to the same image coordinates as a small object closer to the camera (see \figref{fig:scale_fig}). As such, the absolute 3D depth cannot be estimated. The interactions between humans and objects can provide crucial cues to reason about the relative spatial arrangement among them. For example, knowing that two people are riding on the same bike suggests that they should have similar depths
in Fig~\ref{fig:Overview}. This pair-wise interaction cue can be propagated to determine the spatial arrangement of multiple humans and objects together. 
Furthermore, given the fact that a wide range of 2D and 3D supervision exists for human pose, 
we can leverage 3D human pose estimation to further adjust the orientation of the 3D object. For instance, knowing that a person is sitting down on the bench can provide a strong prior to determine the 3D orientation of the bench. Leveraging this requires two important steps: (1) identifying a human and an object that are interacting and (2) defining an objective function to correctly adjust their spatial arrangements.

\begin{figure*}[t]
	\centering
	\includegraphics[width=0.9\textwidth]{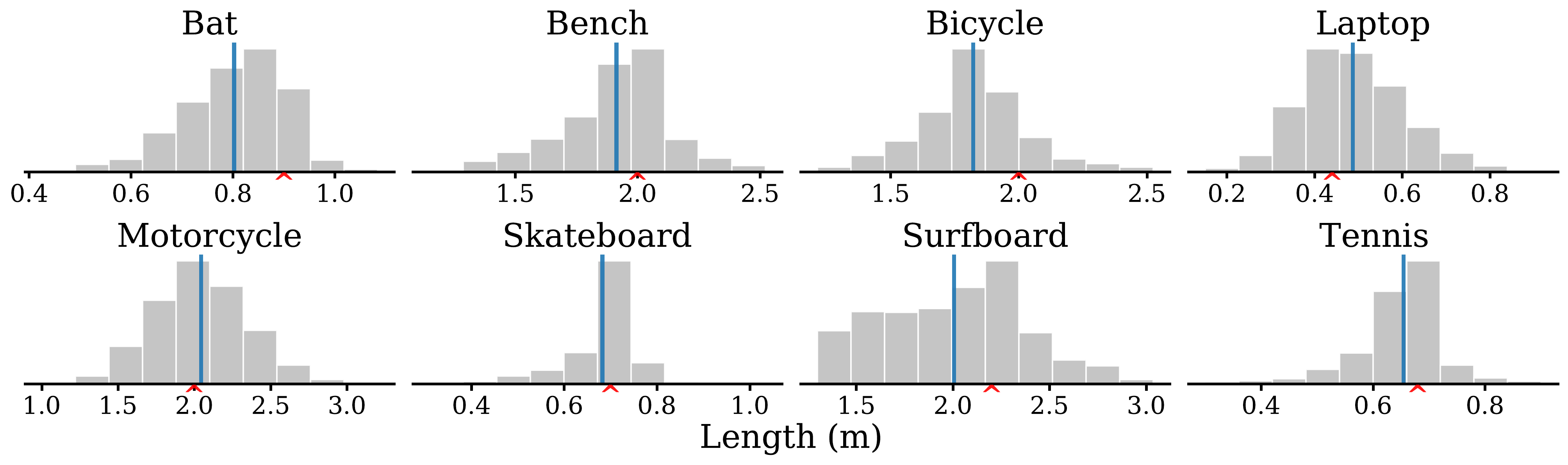}
	
	\caption{\textbf{Learned size distribution from human interaction:} Here, we visualize the distribution of object sizes across the COCO-2017~\cite{lin2014coco} test set at the end of optimization. The red caret denotes the size resulting from the hand-picked scale used for initialization. The blue line denotes the size produced by the empirical mean scale of all category instances at the end of optimization. We then use the empirical mean as the new initialization for intrinsic object scale.
	}
	\label{fig:scales}
\end{figure*}

\paragraph{\textbf{Identifying human-object interaction.}}
We hypothesize that an interacting person and object must be nearby in the world coordinates. 
In our formulation, we solve for the 6-DoF object pose and the intrinsic scale parameter $s^j$ which places the object into world coordinates and imbues the objects with metric size.
We use 3D bounding box overlap between the person and object to determine whether the object is interacting with a person. The size of the per-category 3D bounding box in world coordinates is set larger for larger object categories. See the supplementary materials in \secref{sec:supp} for a full list of 3D box sizes.
Endowing a reasonable initial scale is important for identifying human-object interaction because 
if the object is scaled to be too large or too small in size, it will not be nearby the person. We first initialize the scale using common sense reasoning, via an internet search to find the average size of objects (e.g.~baseball bats and bicycles are $\sim$0.9 meters and $\sim$2 meters long respectively). Through our proposed method, the per-instance intrinsic scales change during optimization. From the final distribution of scales obtained over the test set, we compute the empirical mean scale and repeat this process using this as the new initialization (\figref{fig:scales}).

\paragraph{\textbf{Objective function to optimize 3D spatial arrangements.}}
Our objective includes multiple terms to provide constraints for interacting humans and objects:
\begin{equation}
\label{eq:opt_function}
L = \lambda_1 \loss{occ-sil} + \lambda_2 \loss{interaction} + \lambda_3 \loss{scale} +  \lambda_4 \loss{depth} + \lambda_5 \loss{collision} .
\end{equation}
We optimize \eqref{eq:opt_function} using a gradient-based optimizer~\cite{kingma2014adam} w.r.t.~intrinsic scale $s^i\in\R$ for the $i$-th human and intrinsic scale $s^j\in\R$, rotation $\mathcal{R}^j\in SO(3)$, and translation $\mathbf{t}^j\in \R^3$ for the $j$-th object instance jointly. The object poses are initialized from \secref{sec:nmr_opt}.
$\loss{occ-sil}$ is the same as \eqref{eq:sil} except without the chamfer loss which didn't help during joint optimization. We define the other terms below.

\textbf{Interaction loss}: We first introduce a coarse, instance-level interaction loss to pull the interacting object and person close together:
\begin{equation}
\loss{coarse inter} = \sum_{h\in \mathcal{H}, o\in \mathcal{O}} \mathbbm{1}(h, o)\norm{C(h) - C(o)}_2,
\end{equation}
where $\mathbbm{1}(h, o)$ identifies whether human $h$ and object $o$ are interacting according to the 3D bounding box overlap criteria described before.

Humans generally interact with objects in specific ways.
For example, humans hold tennis rackets by the handle. This can be used as a strong prior for human-object interaction and adjust their spatial arrangement. To do this, we annotate surface regions on the SMPL mesh and on our 3D object meshes where there is likely to be interaction, similar to PROX~\cite{prox}.
These include the hands, feet, and back of a person or the handlebars and seat of a bicycle, as shown in Fig.~\ref{fig:part_labels}. 
To encode spatial priors about human-object interaction (e.g.~people grab bicycle handlebars by the hand and sit on the seat), we enumerate pairs of object and human part regions that interact (see supplementary for a full list).
We incorporate a fine-grained, parts-level interaction loss by using the part-labels (\figref{fig:part_labels}) to pull the interaction regions closer to achieve better alignment:
\begin{equation}
\loss{fine inter} = \sum_{h\in \mathcal{H}, o\in \mathcal{O}} \sum_{\substack{\mathcal{P}_h, \mathcal{P}_o \in \\ \mathcal{P}(h, o)}} \mathbbm{1}(\mathcal{P}_h, \mathcal{P}_o) \norm{C(\mathcal{P}_h) - C(\mathcal{P}_o)}_2,
\end{equation}
where $\mathcal{P}_h$ and $\mathcal{P}_o$ are the interaction regions on the person and object respectively. Note that we define the parts interaction indicator $\mathbbm{1}(\mathcal{P}_h, \mathcal{P}_o)$ using the same criteria as instances, i.e.~3D bounding box overlap. The interactions are recomputed at each iteration. Finally, $\loss{interaction}=\loss{coarse inter} + \loss{fine inter}$.

\textbf{Scale loss}: 
We observe that there is a limit to the variation in size within a category. Thus, we incorporate a Gaussian prior on the intrinsic scales of instances in the same category using a category-specific mean scale:
\begin{equation}
\loss{scale} = \sum_c \sum_{j \in [|\mathcal{O}_c|]}\norm{s^j - \Bar{s}_c}_2.
\end{equation}
We initialize the intrinsic scale of all objects in category $c$ to $\bar{s}_c$. 
The mean object scale $\bar{s}_c$ is initially set using common sense estimates of object size. In \figref{fig:scales}, we visualize the final distribution of object sizes learned for the COCO-2017 \cite{lin2014coco} test set after optimizing for human interaction. We then repeat the process with the empirical mean as a better initialization for $\bar{s}_c$.
We also incorporate the scale loss for the human scales $s^i$ with a mean of 1 (original size) and a small variance.

\textbf{Ordinal Depth loss}: 
The depth ordering inferred from the 3D placement should match that of the image. 
While the correct depth ordering of people and objects would also minimize the occlusion-aware silhouette loss, we posit that the ordinal depth loss introduced in Jiang et al~\cite{jiang2020coherent} can help recover more accurate depth orderings from the modal masks. Using an ordinal depth can give smoother gradients to both the occluder and occluded object.
Formally, for each pair of instances, we compare the pixels at the intersections of the silhouettes with the segmentation mask. If at pixel $p$, instance $i$ is closer than instance $j$ but the segmentation masks at $p$ show $j$ and not $i$, then we apply a ranking loss on the depths of both instances at pixel $p$:
{\small
	\begin{equation}
	\loss{depth} = 
	{\sum_{o^i \in \mathcal{H} \cup \mathcal{O}} \sum_{o^j \in  \mathcal{H} \cup \mathcal{O}} \sum_{\substack{p\in  \text{Sil}(o^i) \\ \cap \text{Sil}(o^j)}} \mathbbm{1}(p, o^i, o^j)
		\log\left(1+\exp(D_{o^j}(p) - D_{o^i}(p))\right)},
	\end{equation}}
where $\text{Sil}(o)$ is the rendered silhouette of instance $o$, $D_o(p)$ is the depth of instance $o$ at pixel $p$, 
and $\mathbbm{1}(p, o^i, o^j)$ is 1 if the segmentation label at pixel $p$ is $o^j$ but $D_{o^i}(p) <  D_{o^j}(p)$.
See \cite{jiang2020coherent} for more details.

\textbf{Collision loss}: Promoting proximity between people and objects can exacerbate the problem of instances occupying the same 3D space. To address this, we penalize poses that would 
human and/or object
interpenetration using the collision loss $\loss{collision}$ introduced in \cite{ballan2012motion,tzionas2016capturing}. We use a GPU implementation based on \cite{SMPL-X:2019} which detects colliding mesh triangles, computes a 3D distance field, and penalizes based on the depth of the penetration. See \cite{SMPL-X:2019} for more details.

%% file: 04_experiments.tex
\begin{figure}
	\centering
	
	\includegraphics[width=0.9\textwidth]{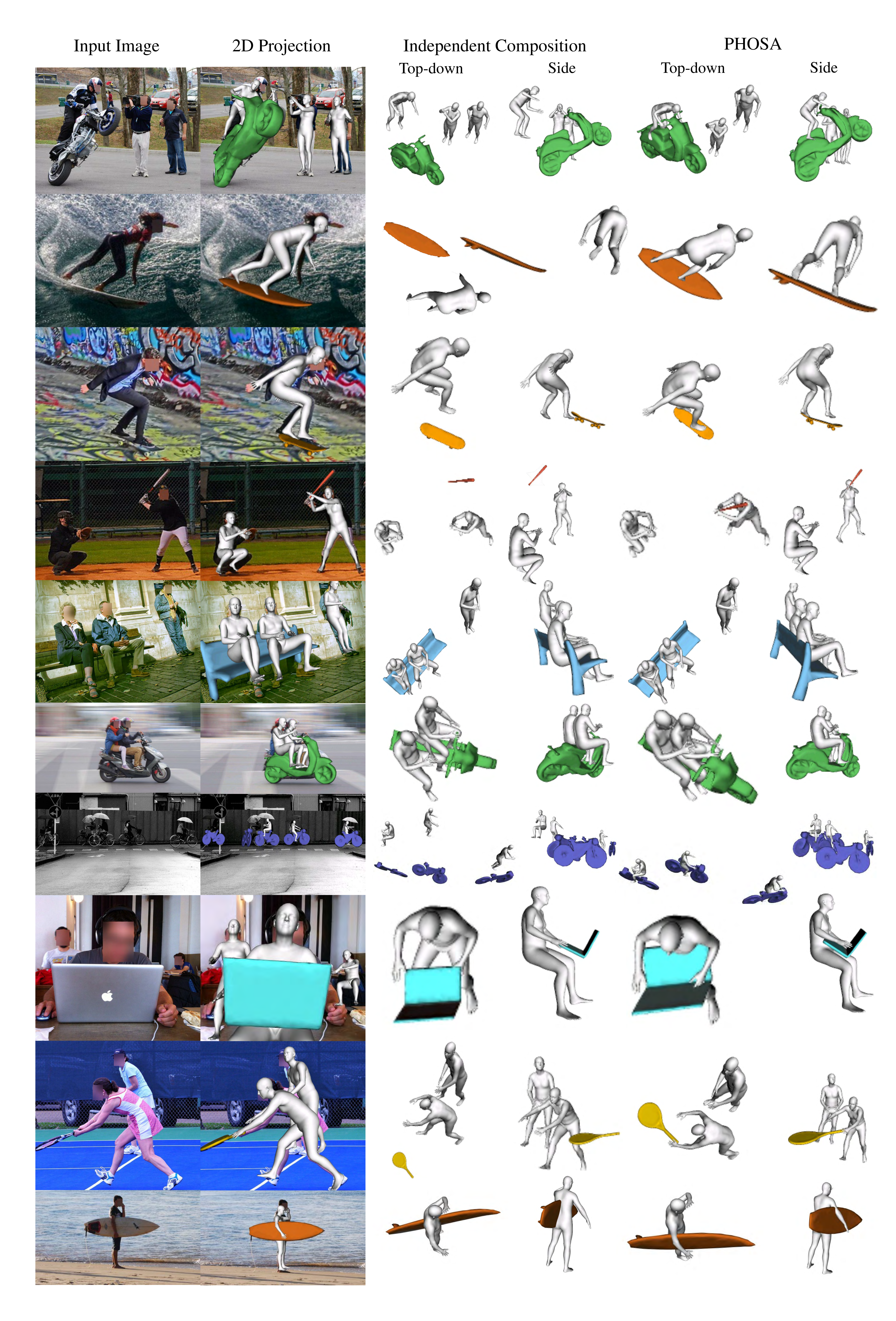}
	\caption{\textbf{Qualitative results of our method on test images from COCO 2017.} Our method, PHOSA, recovers plausible human-object spatial arrangements by explicitly reasoning about human interaction. We evaluate the importance of modeling interaction by comparing with independently estimated human and object poses also using our method (Independent Composition). The intrinsic scale for the independent composition is set to the per-category empirical mean scale learned by our method.}
	\label{fig:main_results}
\end{figure}

\section{Evaluation}
\label{sec:quantitative}

In this section, we provide quantitative and qualitative analysis on the performance of our method on the COCO-2017 \cite{lin2014coco} dataset. We focus our evaluation on 8 categories: baseball bats, benches, bicycles, laptops, motorcycles, skateboards, surfboards, and tennis rackets. 
These categories cover a significant variety in size, shape, and types of interaction with humans.

\subsection{Quantitative Analysis}

\begin{table*}[t]
	\centering
	\resizebox{\textwidth}{!}{
		\footnotesize{
			\begin{tabularx}{\textwidth}{m{25mm}<{\centering}@{}c*9{>{\centering\arraybackslash}X}@{}}
				\toprule
				Ours vs.       & Bat &        Bench & Bike  &  Laptop & Motor. & Skate.\ & Surf. & Tennis & Avg.\\\midrule
				Indep. Comp.&           83 & 74 & 73 & 61 & 74 & 71 & 82 & 82 & 75.0\\
				\cmidrule(lr){1-10}
				No $\loss{occ-sil}$&     79 & 74 & 87 & 76 & 70 & 96 & 80 & 77 & 79.9\\
				No $\loss{interaction}$& 82 & 59 & 57 & 46 & 71 & 71 & 76 & 68 & 66.3 \\
				No $\loss{scale}$&       77 & 49 & 54 & 51 & 54 & 55 & 55 & 56 & 56.4\\
				No $\loss{depth}$&       50 & 55 & 55 & 55 & 52 & 50 & 51 & 50 & 52.3 \\
				No $\loss{collision}$&   52 & 40 & 51 & 51 & 50 & 52 & 50 & 50 & 49.5 \\
				\bottomrule\\
			\end{tabularx}
		}
	}
	\caption{\textbf{Percentage of images for which our proposed method performs better on a subset of COCO 2017 test set.} In this table, we evaluate our approach against an independent composition and ablations of our method. The independent composition estimates human and object pose independently using \secref{sec:anthro_rcnn} and \secref{sec:nmr_opt} and sets the intrinsic scale to the empirical mean category scale learned by our method. 
		The ablations each drop one loss term from our proposed method.
		In each row, we compute the average percentage of images for which our method performs better across a random sample of COCO-2017~\cite{lin2014coco} test set images. A number greater than 50 implies that our proposed method performs better than the independent composition or that the ablated loss term is beneficial for that category.
	}
	\label{tab:interaction}
\end{table*}

Since 3D ground truth annotations for both humans and objects do not exist for in-the-wild images, we used a forced-choice evaluation procedure on COCO-2017~\cite{lin2014coco} images.
To test the contribution of the holistic processing of human and object instances, we evaluate our approach against an ``independent composition," which uses our approach from \secref{sec:anthro_rcnn} and \secref{sec:nmr_opt} to independently estimate the human and object poses.
To make the independent composition competitive, we set the intrinsic scale to be the empirical mean per-category scale learned over the test set by our proposed method in \secref{sec:human_object}. This actually gives the independent composition global information
through human-object interaction. This is the best that can be done without considering all instances holistically.

We collected a random subset of images from the COCO 2017 test set in which at least one person and object overlap in 2D. For each of our object categories, we randomly sample 50 images with at least one instance of that category. For each image, annotators see the independent composition and the result of our proposed method in random order, marking whether our result looks better than, equal to, or worse than the independent composition (see supplementary for screenshots of our annotating tool). We compute the average percentage of images for which our method performs better (treating equal as 50) in \tabref{tab:interaction}. Overall, we find that our method safely outperforms the independent composition.

To evaluate the importance of the individual loss terms, we run an ablative study.
We run the same forced choice test for the full proposed method compared with dropping a single loss term in \tabref{tab:interaction}.
We find that omitting the occlusion-aware silhouette loss and the interaction loss has the most significant effect. Using the silhouette loss during global optimization ensures that the object poses continue to respect image evidence, and the interaction loss encodes the spatial arrangement of the object relative to the person. We did observe that the interaction loss occasionally pulls the object too aggressively toward the person for the laptop category.
The scale loss appears to have a positive effect for most categories. Note that because we initialized the scale to the empirical mean, the effects of the scale loss are not as pronounced as they would be if initialized to something else.
The depth ordering loss gave a moderate boost while the collision loss had a less pronounced effect. We attribute this to the collision loss operating only on the surface triangles and thus prone to getting stuck in local minima in which objects get embedded inside the person (especially for benches).

\subsection{Qualitative Analysis}

In \figref{fig:main_results}, we demonstrate the generality of our approach on the COCO 2017 dataset by reconstructing 
the spatial arrangement of multiple people and objects engaged in a wide range of 3D human-object interactions.
We find that our method works on a variety of everyday objects with which people interact, ranging from handheld objects (baseball bats, tennis rackets, laptops) to full-sized objects (skateboards, bicycles, motorcycles) to large objects (surfboards, benches).
In the middle column of \figref{fig:main_results}, we visualize the spatial arrangement produced by the independent composition introduced in \secref{sec:quantitative}. We find that independently estimating human and object poses is often insufficient for resolving fundamental ambiguities in scale. Explicitly reasoning about human-object interaction produces more realistic spatial arrangements.
Please refer to the supplementary materials in \secref{sec:supp} for discussion of failure modes and significantly more qualitative results.

%% file: 05_discussion.tex
\section{Discussion}

In summary, we have found that 2D and 3D technologies for understanding objects and humans have advanced considerably. Armed with these advances, we believe the time is right to start tackling broader questions of holistic 3D scene-understanding---and moving such questions from the lab to uncontrolled in-the-wild imagery! Our qualitative analysis suggests that 3D human understanding has particularly matured, even for intricate interactions with objects in the wild. Detailed recovery of 3D object shape is still a challenge, as illustrated by our rather impoverished but surprisingly effective exemplar-based shape model. It would be transformative to learn statistical models (a ``SMPL-for-objects"), and we take the first step by learning the intrinsic scale distribution from data.

A number of conclusions somewhat surprised us. First, even though object shape understanding lacks some tools compared to its human counterpart (such as statistical 3D shape models and keypoint detectors), 2D object instance masks combined with a differentiable renderer and a 3D shape library proves to be a rather effective initialization for 3D object understanding. Perhaps even more remarkable is the scalability of such an approach. Adding new objects and defining their modes of interactions is relatively straightforward, because it is far easier to ``paint" annotations on 3D models than annotate individual image instances. Hence 3D shapes provide a convenient coordinate frame for {\em meta}-level supervision. This is dramatically different from the typical supervised pipeline, in which adding a new object category is typically quite involved.

While the ontology of objects will likely be diverse and continue to grow and evolve over time, humans will likely remain a consistent area of intense focus and targeted annotation. Because of this, we believe it will continue to be fruitful to pursue approaches that leverage contextual constraints from humans that act as ``rulers" to help reason about objects. In some sense, this philosophy harkens back to Protagoras's quote from Ancient Greece\,---``man is the measure of all things"!\\

\noindent \textbf{Acknowledgements:} 
We thank Georgia Gkioxari and Shubham Tulsiani for insightful discussion and Victoria Dean and Gengshan Yang for useful feedback. We also thank Senthil Purushwalkam for deadline reminders.
This work was funded in part by the CMU Argo AI Center for Autonomous Vehicle Research.

%% file: 06_appendix.tex
\section{Supplemental Material}
\label{sec:supp}

\begin{figure}
	\centering
	\begin{subfigure}{\textwidth}
		\includegraphics[width=\textwidth]{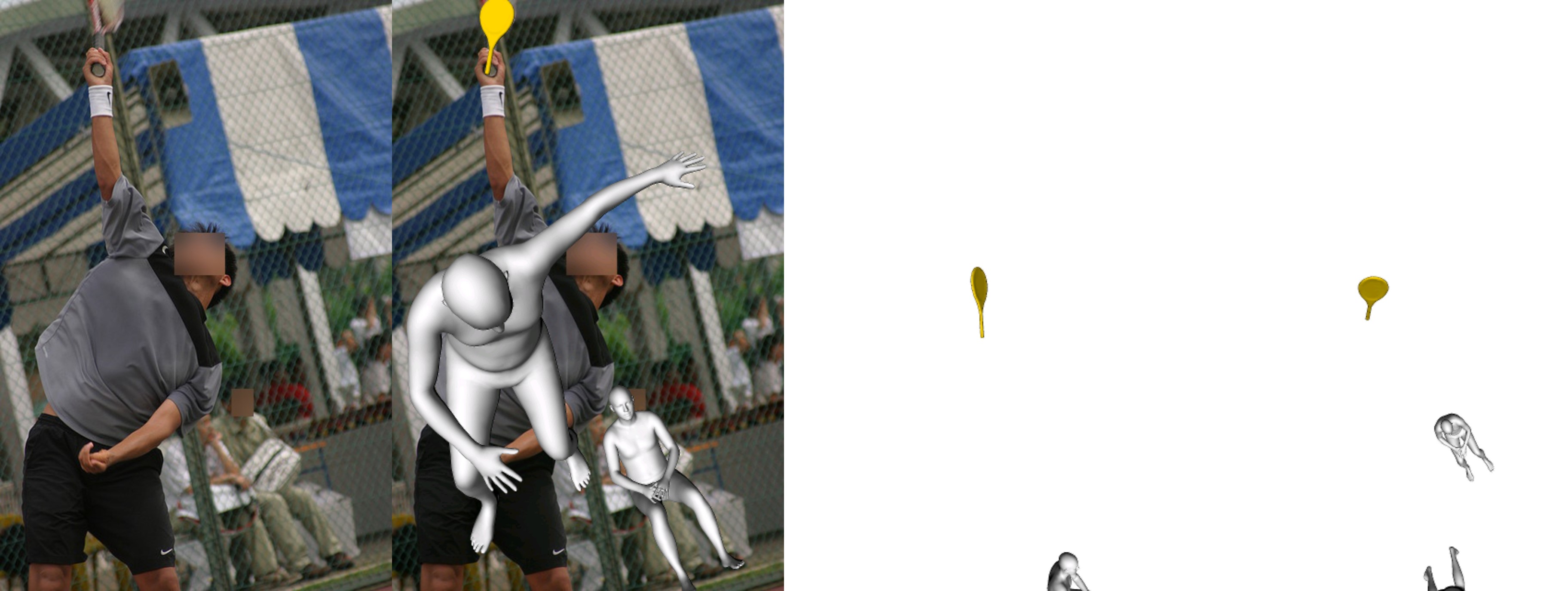}
		\caption{\textbf{Human pose failure.} Our human pose estimator sometimes incorrectly estimates the pose of the person, such as in this challenging snapshot of a tennis player performing a volley. In such cases, it can be difficult to reason properly about human-object interaction since the hand is misplaced and far from the tennis racket.\\
		}
		\label{fig:failure_human_pose}
	\end{subfigure}
	
	\begin{subfigure}{\textwidth}
		\centering
		\includegraphics[width=\textwidth]{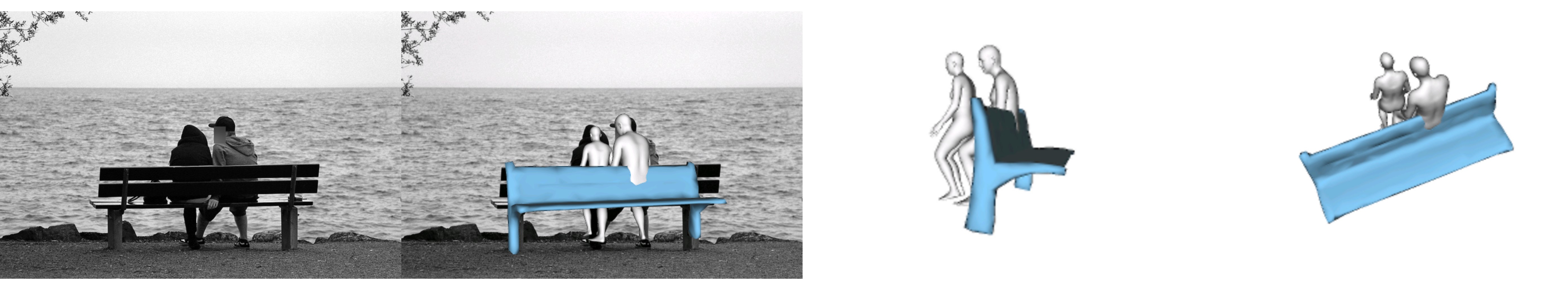}
		\caption{\textbf{Object pose failure.} The predicted masks are sometimes unreliable for estimating the pose of the object. In such cases, it difficult to recover a plausible scene reconstruction.\\
		}
		\label{fig:failure_object_pose}
	\end{subfigure}
	
	\begin{subfigure}{\textwidth}
		\centering
		\includegraphics[width=\textwidth]{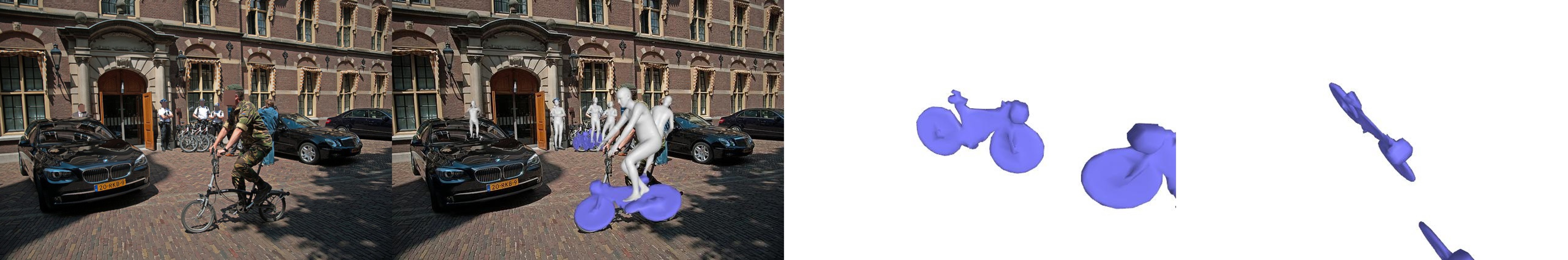}
		\caption{\textbf{Incorrect reasoning about interaction due to scale.} The interaction loss requires a reasonable scale initialization. Sometimes, objects in the real world can fall outside the expected scale distribution, such as in the case of this small bicycle.
		}
		\label{fig:failure_interaction}
	\end{subfigure}
	\caption{\textbf{Failure modes.} In this figure, we describe a few failure modes of our method.}
	\label{fig:failure_modes}
\end{figure}

In this section, we describe implementation details in \secref{sec:implementation_details} and the mesh processing pipeline in \secref{sec:mesh_processing}. We also include more qualitative results in \secref{sec:more_qualitative} and describe a few failure modes in \figref{fig:failure_modes}.

\subsection{Implementation details}
\label{sec:implementation_details}

We represent rotations for the object poses using the 6-DoF rotation representation introduced in \cite{zhou2019continuity}. We optimize the the occlusion aware silhouette loss using the ADAM optimizer~\cite{kingma2014adam} with learning rate 1e-3 for 100 iterations. We compute the edge maps $E(M)$ using $\text{MaxPool}(M) - M$ with a filter size of 7. Since the occlusion-aware silhouette loss is susceptible to getting stuck in local minima, we initialize with 10,000 randomly generated rotations and select the pose that produces the lowest loss value. For some categories (bicycle, bench, motorcycle), we found it beneficial to bias the sampling toward upright poses (elevation between -30 and 30 degrees, azimuth between 0 and 360 degrees).

We jointly optimize the 3D spatial arrangement loss \eqref{eq:opt_function} using ADAM with learning rate 1e-3 for 400 iterations. The trainable parameters are intrinsic scale $s^i\in\R$ for the $i$-th human and intrinsic scale $s^j\in\R$, rotation $\mathcal{R}^j\in SO(3)$, and translation $\mathbf{t}^j\in \R^3$ for the $j$-th object instance. The loss weights $\lambda_i$ are tuned qualitatively on the COCO-2017 val set. We initialized the optimization with the human poses estimated using \cite{joo2020exemplar} and the best object pose estimated in \secref{sec:nmr_opt} per object instance. To improve computational speed, we downsample the SMPL human meshes to 502 vertices and 1000 faces when computing losses.

A list of interaction parts pairs can be found in \tabref{tab:parts}, and an enumeration of the sizes of the 3D bounding boxes used to compute the interaction losses can be found in \tabref{tab:bbox_sizes}.

\begin{table}[h]
	\centering
	\footnotesize{
		\begin{tabular}{ll}
			\toprule
			Category & Part Pairs (Object Part, Human Part)\\\midrule
			Bat & (Handle, L Palm), (Handle, R Palm)\\
			Bench & (Seat, Butt), (Seat Back, Back)\\
			Bicycle & (Seat, Butt), (Handlebars, L Palm), (Handlebars, R Hand)\\
			Laptop & (Laptop, L Palm), (Laptop, R Palm)\\
			Motorcycle & (Seat, Butt), (Handlebars, L Palm), (Handlebars, R Palm)\\
			Skateboard & (Skateboard, L Foot), (Skateboard, R Foot)\\
			Surfboard & (Surfboard, L Foot), (Surfboard, R Foot), (Surfboard, L Palm)\\
			& (Surfboard, R Palm)\\
			Tennis Racket & (Handle, L Palm), (Handle, R Palm)\\
			\bottomrule\\
		\end{tabular}
	}
	\caption{\textbf{List of Parts Pairs used per category.} Each parts pair consists of a part of an object and a part of the human body. These parts pairs are used to assign human-object interactions.}
	\label{tab:parts}
\end{table}

\begin{table}[h]
	\centering
	\footnotesize{
		\begin{tabular}{lccc}
			\toprule
			Category & XY (Coarse) & XY (Fine) & Z Depth\\\midrule
			Bat & 0.5 & 2.5 & 5\\
			Bench & 0.3 & 0.5 & 10\\
			Bicycle & 0 & 0.7 & 4\\
			Laptop & 0.2 & 0 & 2.5\\
			Motorcycle & 0 & 0.7 & 5\\
			Skateboard & 0 & 0.5 & 10\\
			Surfboard & 0.8 & 0.2 & 50\\
			Tennis Racket & 0.4 & 2 & 5\\
			\bottomrule\\
		\end{tabular}
	}
	\caption{\textbf{Size of 3D Bounding Boxes.} To determine whether to apply the coarse interaction loss, we take the bounding box of the object and the bounding box of the person and expand each by the coarse expansion factor (Column 2). If the expanded bounding boxes overlap and the difference in the depths of the person and object is less than the depth threshold (Column 4), then we consider the person and object to be interacting. To determine whether to apply the fine interaction loss, we similarly take the bounding boxes corresponding to the object part and person part, expand the bounding boxes by the fine expansion factor (Column 3), and check for overlap. If the expanded bounding boxes overlap and the difference in depths of the parts is less than the depth threshold, then we consider the person part and object part to be interacting.}
	\label{tab:bbox_sizes}
\end{table}

\subsection{Pre-processing Mesh Models}
\label{sec:mesh_processing}
\begin{figure}
	\centering
	\includegraphics[width=0.9\textwidth]{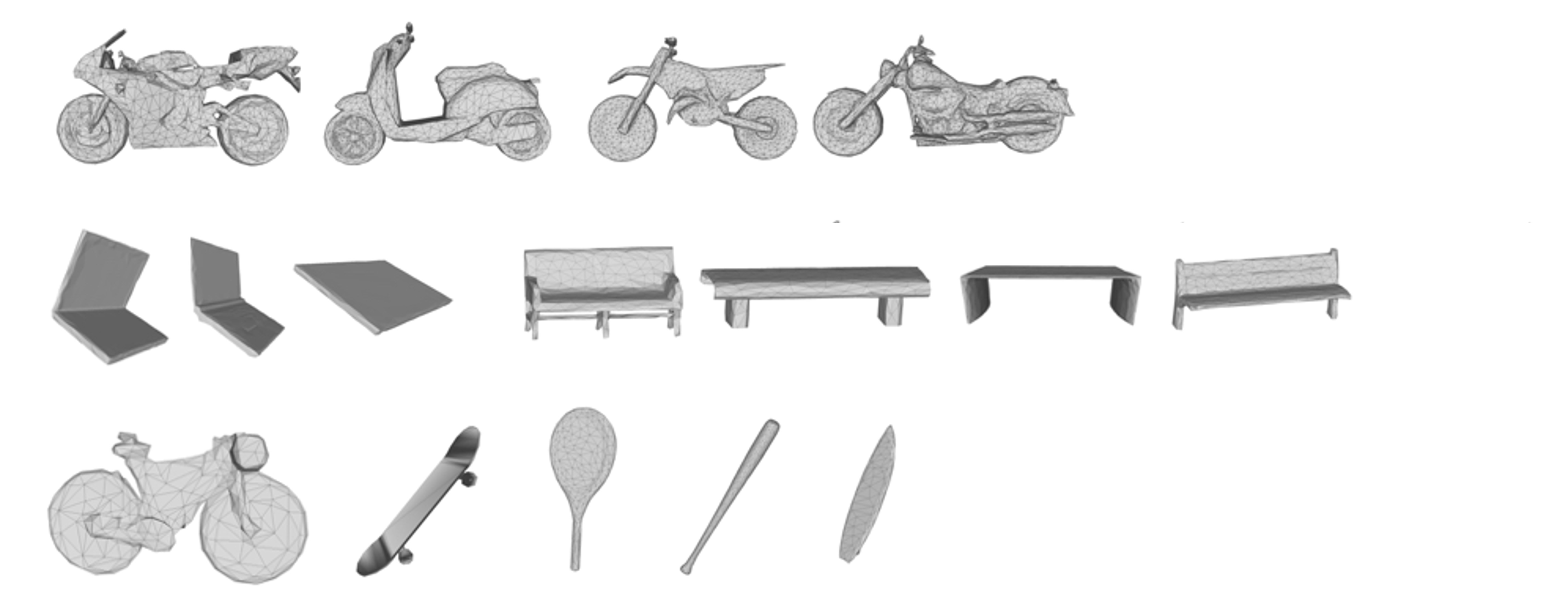}
	\caption{\textbf{Mesh models from various 3D object categories.} Here, we show all the mesh models that we used. \textbf{First row:} Motorcycle. \textbf{Third Row:} Laptop, Bench. \textbf{Fourth Row:} Bicycle, Skateboard, Tennis Racket, Baseball Bat, Surfboard.}
	\label{fig:mesh_instances}
\end{figure}

We show all mesh instances that we built for each 3D category in \figref{fig:mesh_instances}. To better cover the shape variation or articulation within an object category, we use multiple mesh models for a few object categories (e.g., motorcycle, bench, and laptop). All the meshes are pre-processed to be watertight and are simplified with a low number of faces and uniform face size, to make the the optimization more efficient. For the pre-processing, we first fill in the holes of the raw mesh models (e.g. the holes in the wheels or tennis racket) to make the projection of the 3D models consistent with the silhouettes obtained by the instance segmentation algorithm~\cite{kirillov2019pointrend}. Then, we perform a TSDF fusion approach~\cite{Stutz2018ARXIV} that converts the raw meshes to be watertight and simplified. Finally, we reduce the number of mesh vertices using MeshLab~\cite{meshlab}.

\begin{figure}[t]
	\centering
	\fbox{\includegraphics[width=\textwidth]{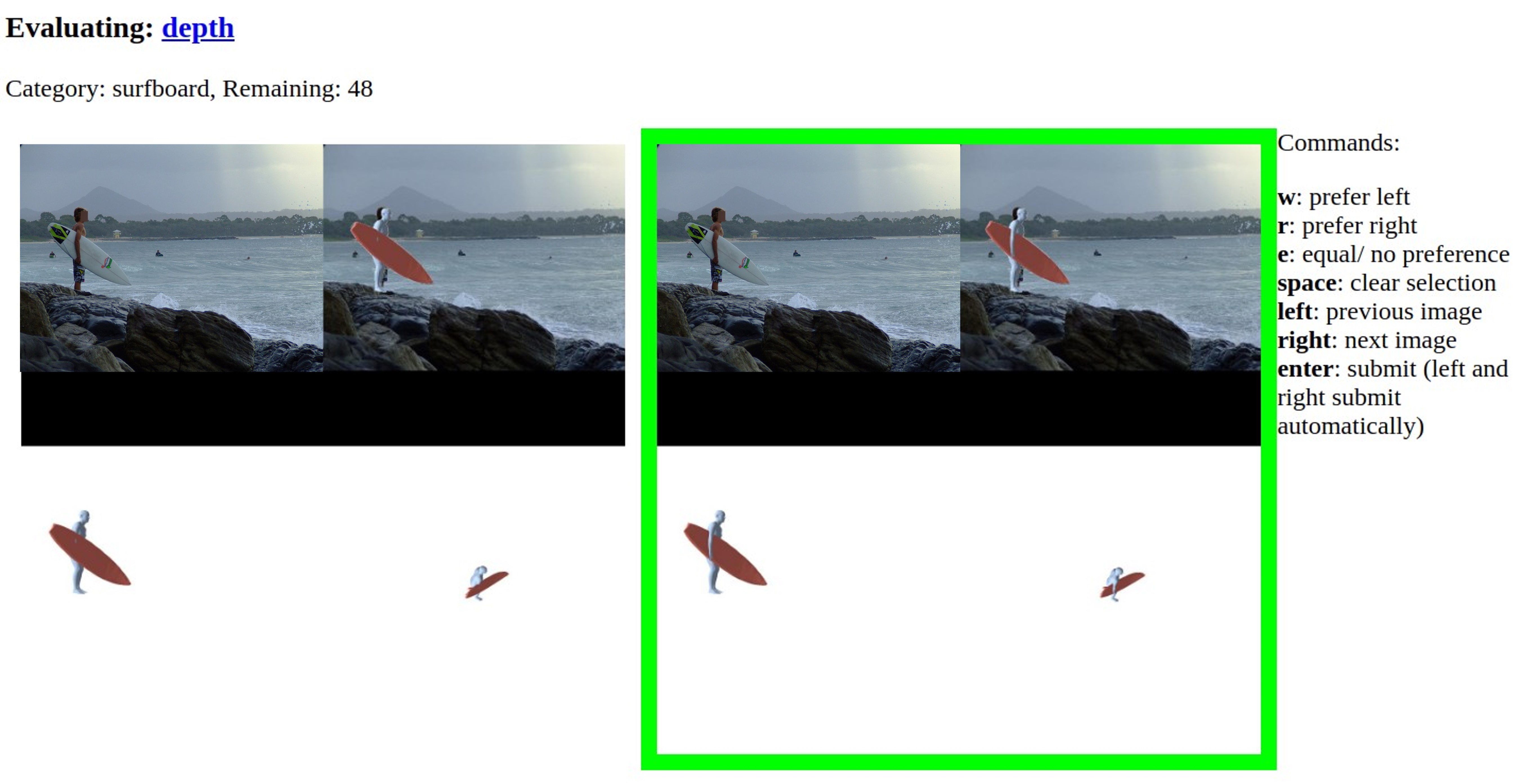}}
	\caption{\textbf{Screenshot from our comparison evaluation test interface.} Annotators were asked to evaluate which 3D arrangement looks more accurate, in this case, without and with the depth ordering loss for a picture of a person with a surfboard. Clockwise from top-left: original image, image with rendered projection, top-down view, frontal view. The arrangement highlighted in green is the one that is selected.}
	\label{fig:evaluation_interface}
\end{figure}

\subsection{More Qualitative Results}
\label{sec:more_qualitative}

\begin{figure}[t]
	\centering
	\includegraphics[width=\textwidth]{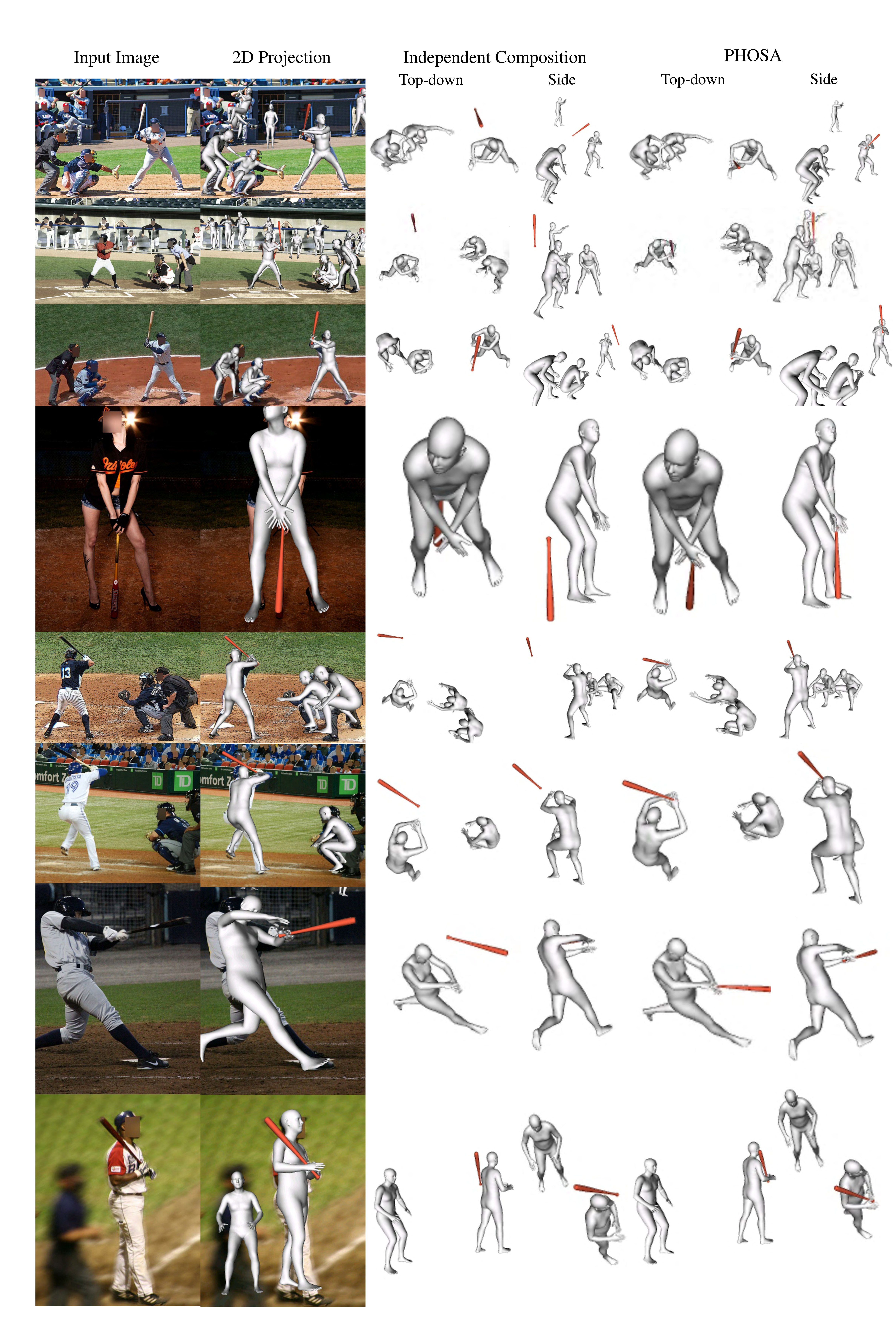}
	\caption{\textbf{Our output on COCO images with baseball bats.}}
	\label{fig:bat}
\end{figure}

\begin{figure}[t]
	\centering
	\includegraphics[width=\textwidth]{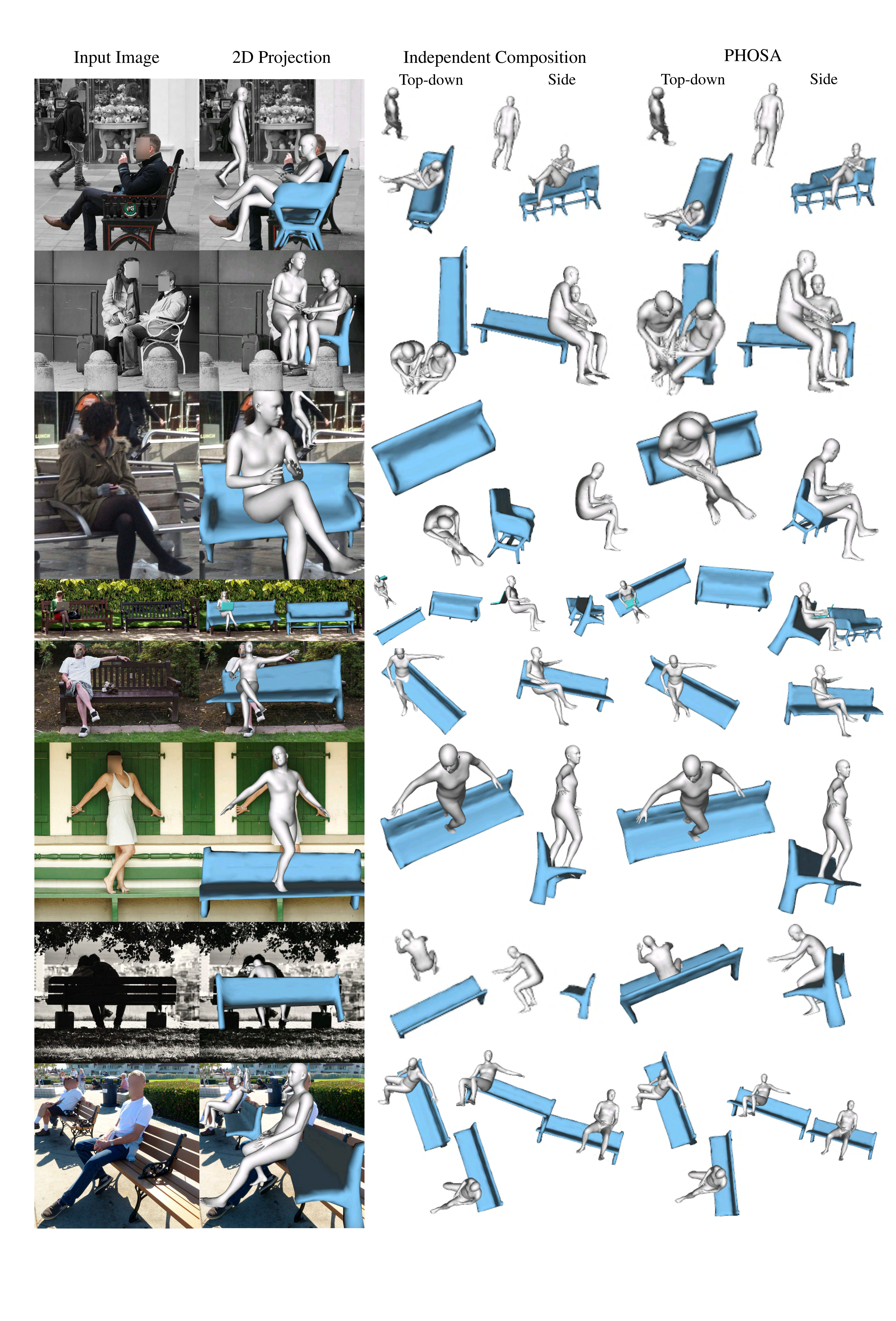}
	\caption{\textbf{Our output on COCO images with benches.}}
	\label{fig:bench}
\end{figure}

\begin{figure}[t]
	\centering
	\includegraphics[width=\textwidth]{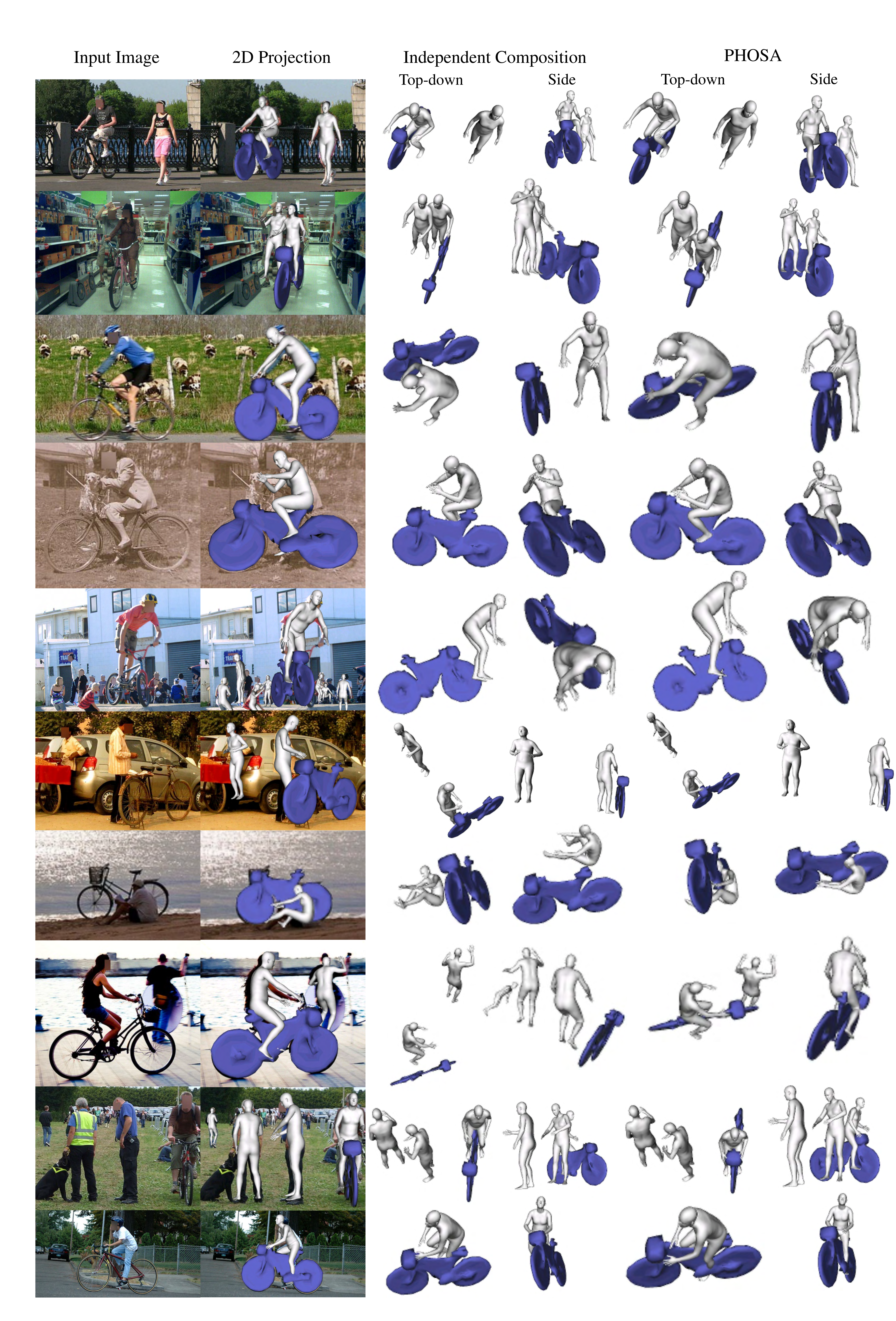}
	\caption{\textbf{Our output on COCO images with bicycles.}}
	\label{fig:bicycle}
\end{figure}

\begin{figure}[t]
	\centering
	\includegraphics[width=\textwidth]{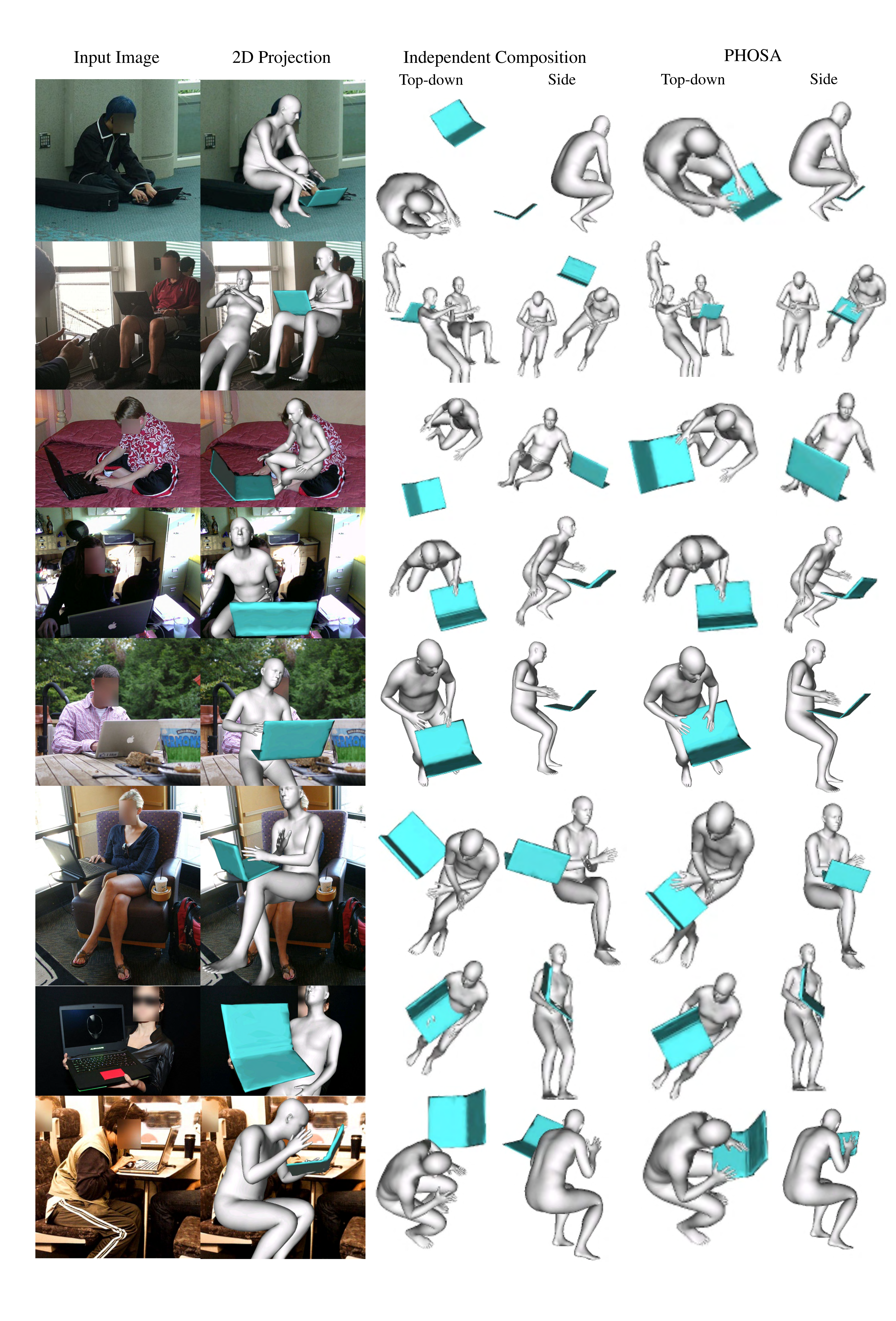}
	\caption{\textbf{Our output on COCO images with laptops.}}
	\label{fig:laptop}
\end{figure}

\begin{figure}[t]
	\centering
	\includegraphics[width=\textwidth]{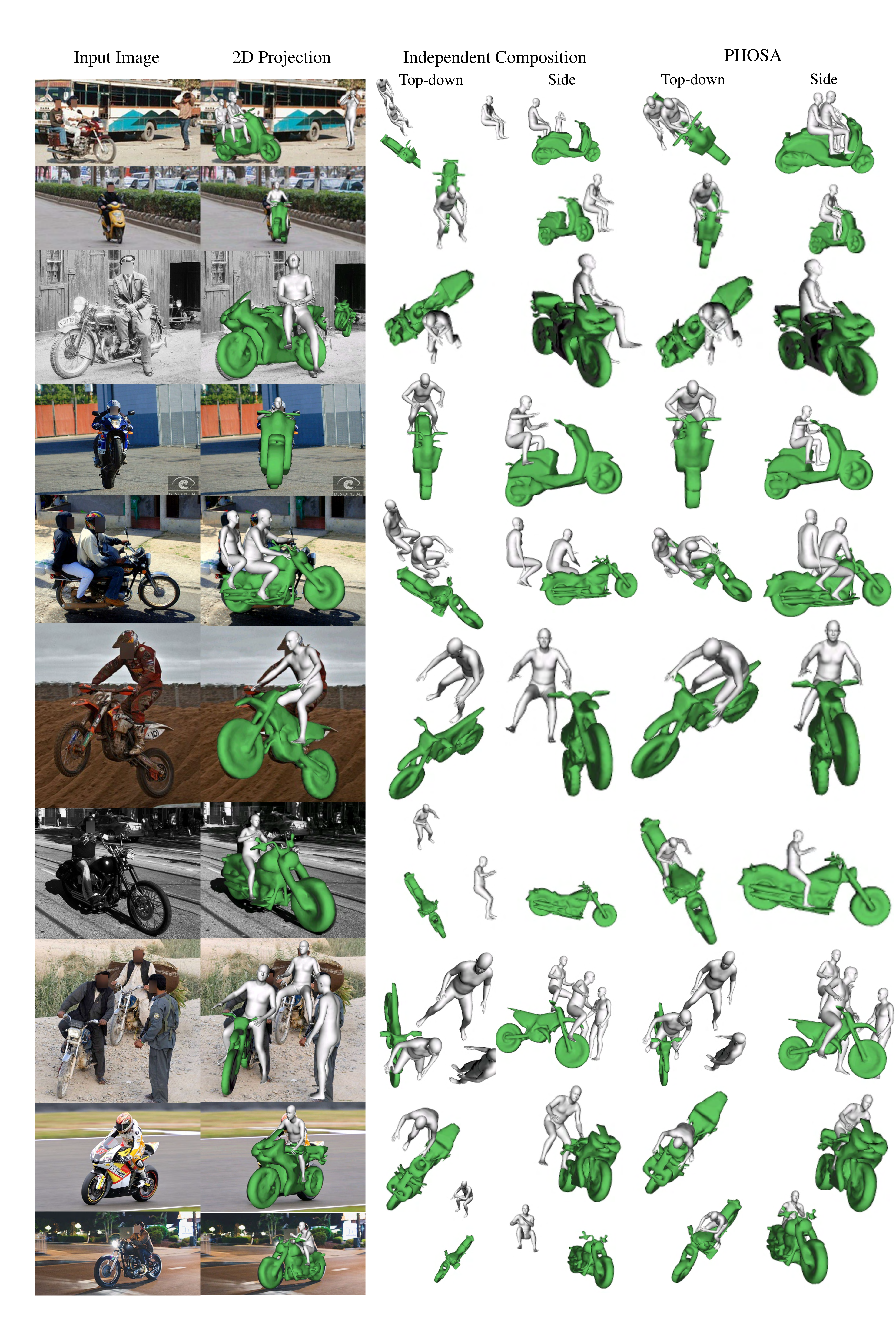}
	\caption{\textbf{Our output on COCO images with motorcycles.}}
	\label{fig:motorcycle}
\end{figure}

\begin{figure}[t]
	\centering
	\includegraphics[width=\textwidth]{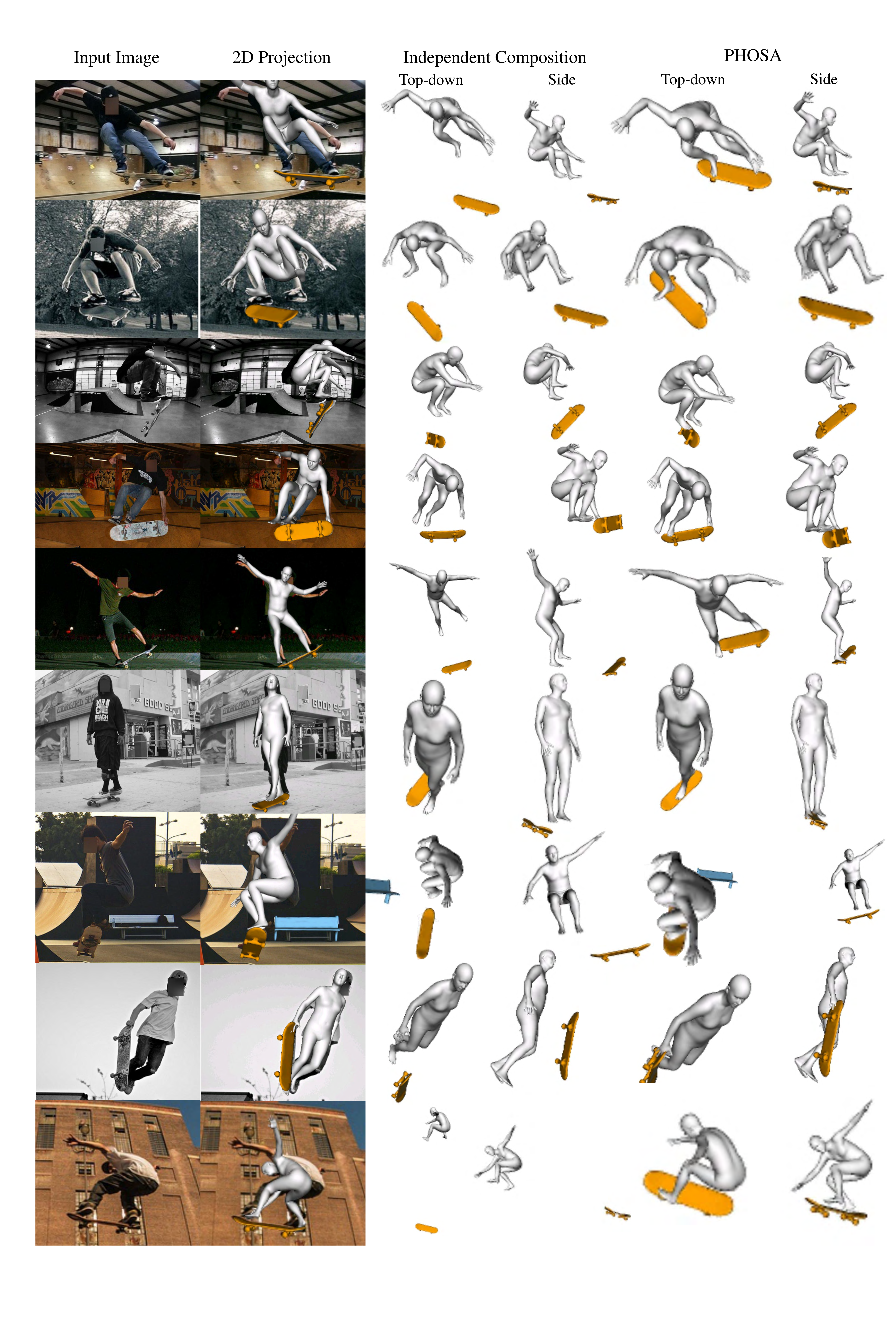}
	\caption{\textbf{Our output on COCO images with skateboards.}}
	\label{fig:skateboard}
\end{figure}

\begin{figure}[t]
	\centering
	\includegraphics[width=\textwidth]{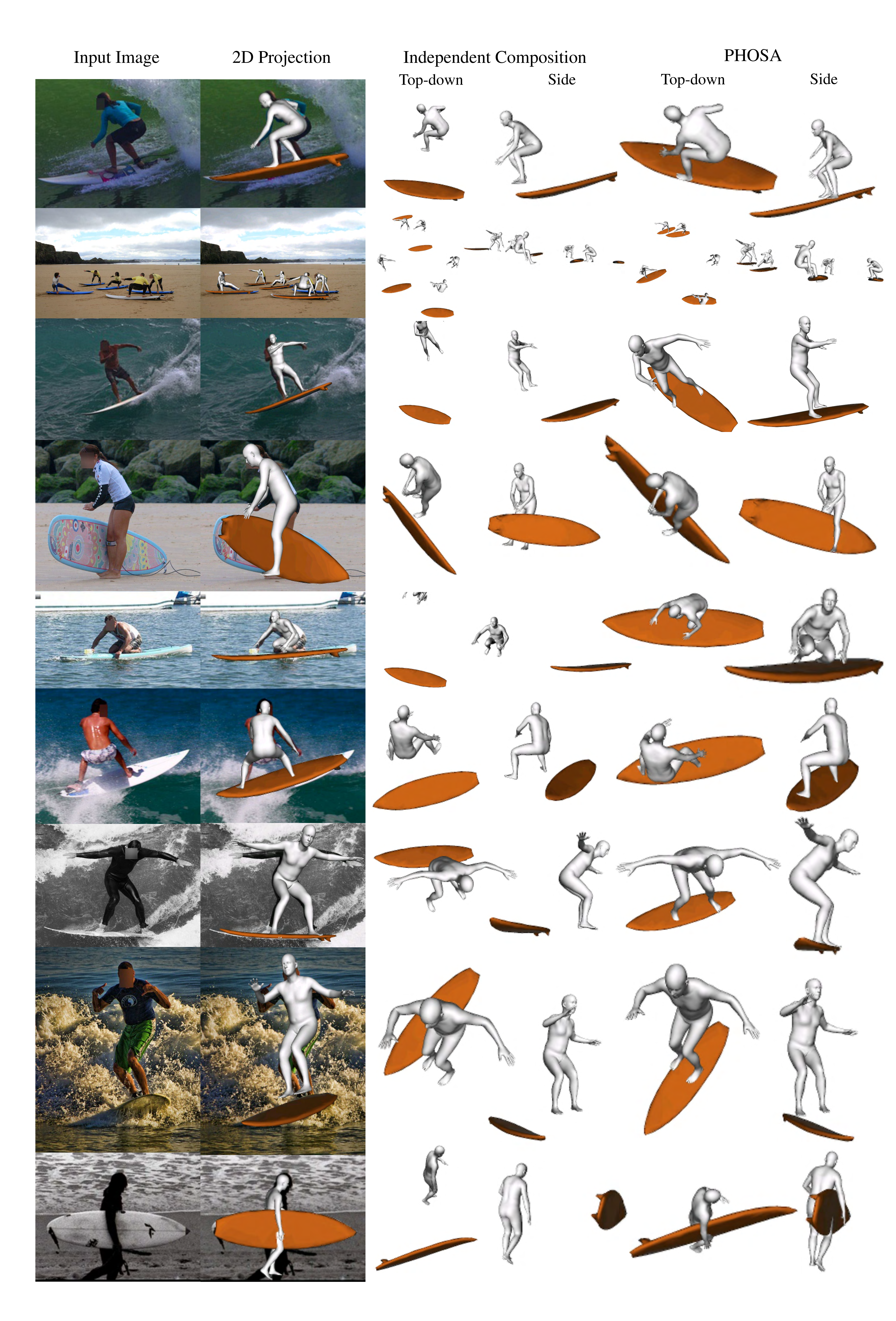}
	\caption{\textbf{Our output on COCO images with surfboards.}}
	\label{fig:surfboard}
\end{figure}

\begin{figure}[t]
	\centering
	\includegraphics[width=\textwidth]{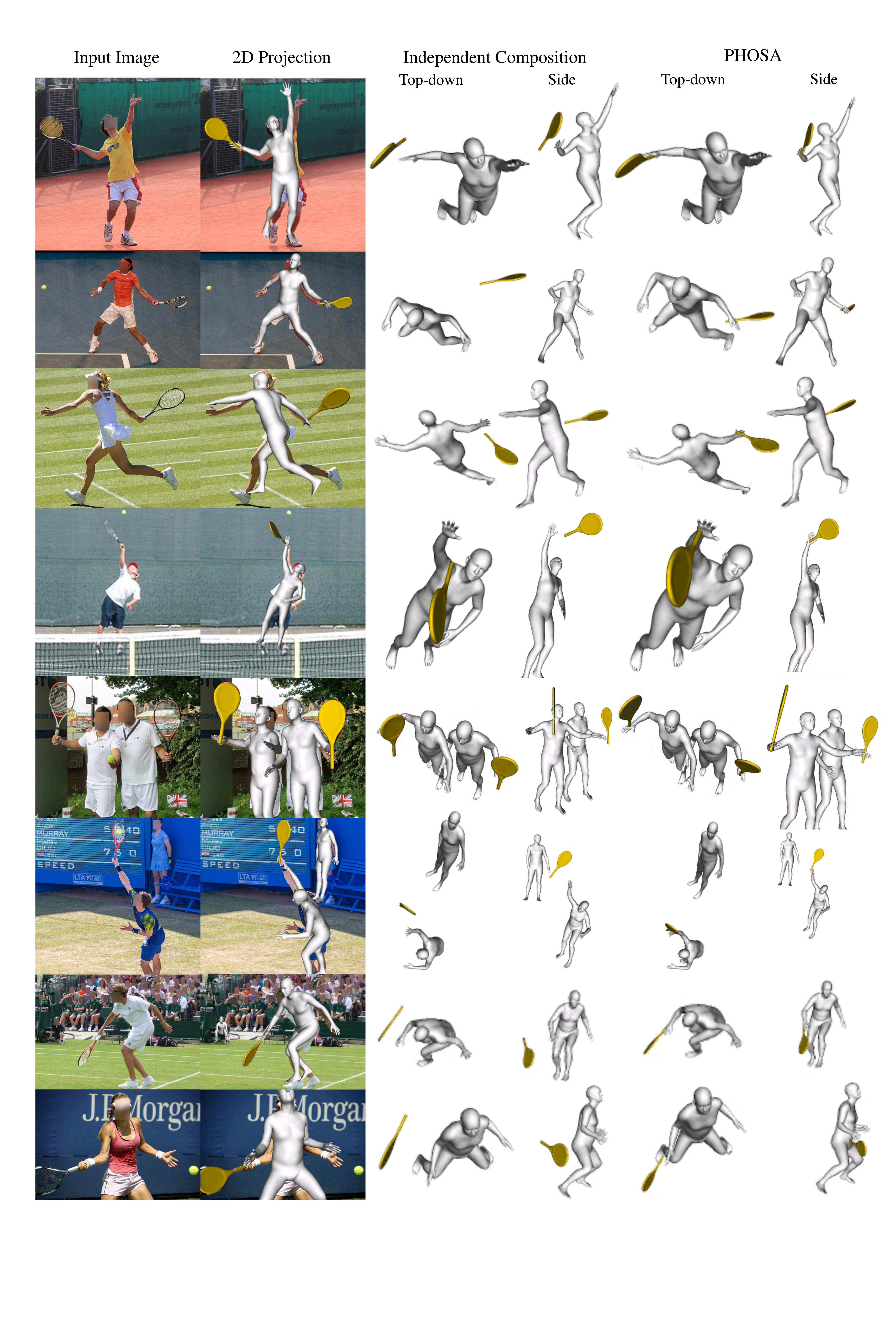}
	\caption{\textbf{Our output on COCO images with tennis rackets.}}
	\label{fig:tennis}
\end{figure}

We show results on a large number of COCO images (test set) for each category evaluated in the main paper: baseball bats (\figref{fig:bat}), benches (\figref{fig:bench}), bicycles (\figref{fig:bicycle}), laptops (\figref{fig:laptop}), motorcycles (\figref{fig:motorcycle}), skateboards (\figref{fig:skateboard}), surfboards (\figref{fig:surfboard}), and tennis rackets (\figref{fig:tennis}). 

